\documentclass{article}
\usepackage[preprint]{neurips_2025}
\usepackage[T1]{fontenc}

\usepackage[protrusion=false]{microtype}\usepackage{graphicx,booktabs,amsmath,amssymb,amsthm,xcolor,enumitem,tabularx,listings,multirow}
\usepackage{caption}
\usepackage{needspace}
\usepackage{placeins}
\usepackage{hyperref}
\hypersetup{colorlinks=true,linkcolor=blue!50!black,citecolor=blue!50!black,urlcolor=blue!50!black,pdftitle={Climbing the SkyLadder: A Certified Lower Bound of 25 for Kochen-Specker Sets in R3},pdfauthor={Zhengyu Li, Vijay Ganesh}}
\usepackage{tikz}
\usetikzlibrary{positioning,arrows.meta,shapes.geometric,fit,backgrounds,calc,decorations.pathreplacing}
\definecolor{agentc}{HTML}{F6D3B3}\definecolor{toolc}{HTML}{C9DDF0}\definecolor{trustc}{HTML}{CDE8C9}\definecolor{humanc}{HTML}{E3E3E3}
\definecolor{agentd}{HTML}{C0661F}\definecolor{trustd}{HTML}{3D8B37}
\tikzset{fbox/.style={draw, rounded corners=3pt, align=center, font=\footnotesize, inner sep=3pt},
  agentbox/.style={fbox, fill=agentc}, toolbox/.style={fbox, fill=toolc}, trustbox/.style={fbox, fill=trustc},
  humanbox/.style={fbox, fill=humanc}, plainbox/.style={fbox, fill=white},
  farr/.style={-{Stealth[length=2mm]}, thick}, flab/.style={font=\scriptsize, align=center}}

\setlist{itemsep=1pt,topsep=3pt}
\newtheorem{theorem}{Theorem}\newtheorem{definition}[theorem]{Definition}
\DeclareUnicodeCharacter{211D}{\ensuremath{\mathbb{R}}}
\DeclareUnicodeCharacter{2115}{\ensuremath{\mathbb{N}}}
\makeatletter\renewcommand{\@noticestring}{Preprint.}\makeatother
\newcommand{\cpu}{\textsc{cpu}}
\newcommand{\R}{\mathbb{R}}
\newcommand{\lean}[1]{\texttt{#1}}
\newcommand{\mc}[1]{\begin{tabular}[b]{@{}c@{}}#1\end{tabular}}
\newcommand{\method}{SkyLadder}
\definecolor{ours}{HTML}{D0612A}\definecolor{hum}{HTML}{7A7A7A}
\lstdefinelanguage{lean4}{morekeywords={def,theorem,abbrev,namespace,end,import,fun,let,by,structure,where},
  sensitive=true,morecomment=[l]{--},morecomment=[s]{/-}{-/},morestring=[b]"}
\title{Climbing the \method: A Certified Lower Bound of 25 for Kochen--Specker Sets in $\R^3$}
\author{%
  Zhengyu Li\\
  Georgia Institute of Technology\\
  \texttt{brian.li@gatech.edu}
  \And
  Vijay Ganesh\\
  Georgia Institute of Technology\\
  \texttt{vganesh45@gatech.edu}
}

\begin{document}
\maketitle

\begin{abstract}
Can an AI agent find \emph{new} techniques for a search problem that experts already optimised? Can we trust a ``no'' answer that the agent's own code produced?
We introduce \method, a protocol for agent-driven research on exact search problems. The agent climbs the problem one size at a time. It develops techniques motivated by measured bottlenecks and records both improvements and dead ends. Certification requires complete coverage by certificate checkers that Lean~4 proves sound.
We apply \method\ to the minimum Kochen--Specker (KS) problem in $\R^3$, a question from the foundations of quantum mechanics. The best lower bound was 24 vectors. It came from two independent SAT-based searches of about 12{,}000 \cpu-hours each.
On one 8-core laptop, the agent's pipeline reproduces every published order from 17 to 23 and matches the published candidate counts exactly. At order 23 it is $2.5\times10^{4}$ times faster than the state of the art. Most of the gain comes from one reformulation: the search runs over \emph{basis structures} instead of vector graphs.
The pipeline resolves the next open case: \emph{every KS set in $\R^3$ has at least 25 vectors}. The certificate checks establish that the enumeration is complete. They cover 2.2 million final basis structures across orders 1--24 and validate 1{,}027 real-algebraic non-embeddability certificates. The retained verification records account for all 1{,}748{,}733 required order-24 structures, including supplementary and cube-split checks.
Lean proves that the checkers are sound. The trusted base also includes the mathematical specification, Lean and Mathlib, compiled checker execution, certificate-file handling, and assembly of per-process logs. Under this explicit trusted base, the result does not depend on the correctness of the search tools.
An ablation and a prior-art audit separate the new techniques from those that the agent rediscovered.
\end{abstract}

\section{Introduction}
\label{sec:intro}

AI systems now find new mathematical \emph{constructions}: a program search proposes an object, and an evaluator scores it~\citep{romeraparedes2024,novikov2025,yuksekgonul2026}. Many open questions in combinatorics have the opposite shape. Their answer is that something does \emph{not} exist. The proof is an exhaustive search. Progress comes from techniques that make the search smaller, not from one clever object. Here it is harder both to measure and to trust the contribution of an agent.

\begin{center}
\fbox{\begin{minipage}{0.94\linewidth}\small
\textbf{Key question.} Can an AI agent advance an exhaustive search that experts already optimised? We require two properties. (i)~The method attributes every gain to an identifiable technique and says which techniques are new. (ii)~The non-existence result does not depend on trust in the agent's search code.
\end{minipage}}
\end{center}

\paragraph{The test case.} The Kochen--Specker theorem~\citep{kochen1967} shows that quantum mechanics in dimension at least three admits no non-contextual hidden-variable model. In $\R^3$ its proof needs a finite set of vectors that cannot be ``coloured'' consistently (Section~\ref{sec:bg}). The smallest known such set has 31 vectors and is due to Conway and Kochen~\citep{peres1993}. Lower bounds come from exhaustive search: 18~\mbox{\citep{arends2011}}, 22~\mbox{\citep{uijlen2016}}, and 24. Two methods reached 24 independently: SAT with computer algebra (SAT+CAS)~\citep{li2024} and SAT Modulo Symmetries (SMS)~\citep{kirchweger2023}. Each of these two searches spent about $4.3\times10^{7}$ \cpu-seconds on order 23 alone. From order 20 on, the cost grew more than $10\times$ per order. The established interval was $[24,31]$, so order 24 was the next open case.

\paragraph{Two obstacles.} First, ``the agent beat the baseline'' says little unless we know which ideas are new and which re-implement the literature. Second, consider code from an agent that already concluded that the answer is ``no''. That code is exactly the code one should not trust. Agent-reported success is known to be unreliable~\citep{tang2025airesearcher}.

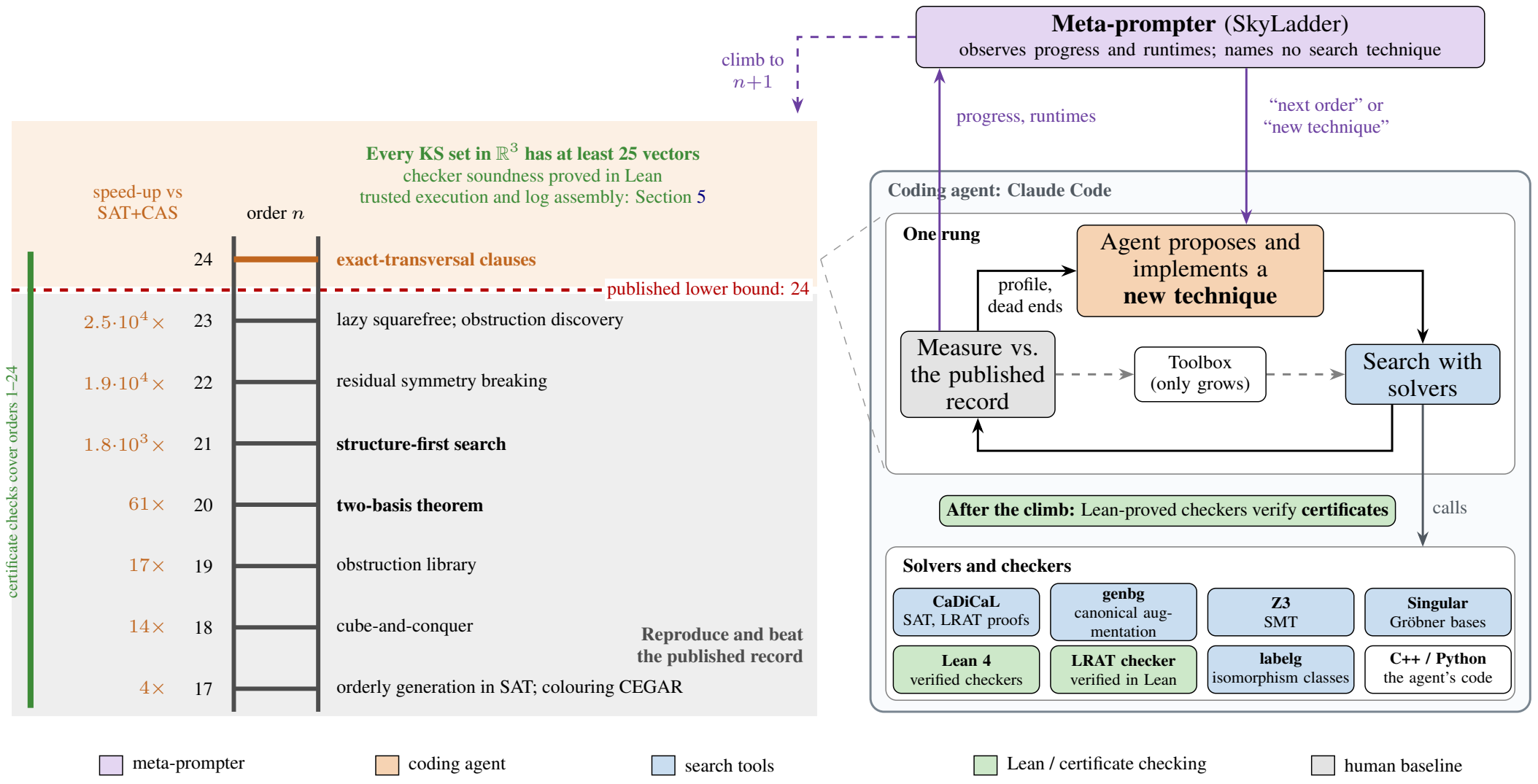
\begin{figure}[t]
\centering
\resizebox{\linewidth}{!}{
\begin{tikzpicture}[x=1cm,y=1cm]
  \fill[humanc!60] (-2.9,-0.35) rectangle (7.6,5.15);
  \fill[agentc!35] (-2.9,5.25) rectangle (7.6,7.4);
  \node[font=\scriptsize\bfseries, anchor=east, align=right, text=black!70] at (7.55,0.55) {Reproduce and beat\\the published record};
  \draw[red!70!black, very thick, dashed] (-2.9,5.2) -- (7.6,5.2);
  \node[font=\scriptsize, red!70!black, fill=white, inner sep=1pt, anchor=east] at (7.55,5.2) {published lower bound: 24};
  \foreach \n in {17,...,24} {
    \pgfmathsetmacro{\y}{(\n-17)*0.8}
    \node[font=\scriptsize, anchor=east] at (-0.15,\y) {\n};
  }
  \draw[line width=1.6pt, black!70] (0,-0.3) -- (0,5.9);
  \draw[line width=1.6pt, black!70] (1.1,-0.3) -- (1.1,5.9);
  \foreach \n in {17,...,23} { \pgfmathsetmacro{\y}{(\n-17)*0.8} \draw[line width=1.6pt, black!70] (0,\y) -- (1.1,\y); }
  \draw[line width=2.2pt, agentd] (0,5.6) -- (1.1,5.6);
  \node[font=\scriptsize, anchor=south] at (0.55,6.0) {order $n$};
  \node[font=\scriptsize, anchor=south, align=center, text=agentd] at (-1.25,6.0) {speed-up vs\\SAT+CAS};
  \foreach \n/\g in {17/4,18/14,19/17,20/61,21/1.8{\cdot}10^3,22/1.9{\cdot}10^4,23/2.5{\cdot}10^4} {
    \pgfmathsetmacro{\y}{(\n-17)*0.8}
    \node[font=\scriptsize, text=agentd, anchor=east] at (-0.75,\y) {$\g\times$};
  }
  \draw[trustd, line width=2.2pt] (-2.65,-0.25) -- (-2.65,5.7);
  \node[font=\tiny, text=trustd, rotate=90, anchor=south] at (-2.7,2.7) {certificate checks cover orders 1--24};
  \tikzset{ann/.style={font=\scriptsize, anchor=west, inner sep=1pt}}
  \node[ann] at (1.3,0.0) {orderly generation in SAT; colouring CEGAR};
  \node[ann] at (1.3,0.8) {cube-and-conquer};
  \node[ann] at (1.3,1.6) {obstruction library};
  \node[ann] at (1.3,2.4) {\textbf{two-basis theorem}};
  \node[ann] at (1.3,3.2) {\textbf{structure-first search}};
  \node[ann] at (1.3,4.0) {residual symmetry breaking};
  \node[ann] at (1.3,4.8) {lazy squarefree; obstruction discovery};
  \node[ann, text=agentd] at (1.3,5.6) {\textbf{exact-transversal clauses}};
  \node[font=\scriptsize, text=trustd, align=center] at (3.9,6.7) {\textbf{Every KS set in $\R^3$ has at least 25 vectors}\\checker soundness proved in Lean\\trusted execution and log assembly: Section~\ref{sec:cert}};
  \definecolor{metac}{HTML}{E4D6F2}\definecolor{metad}{HTML}{6B3FA0}
  \node[fbox, fill=metac, text width=7.2cm] (M) at (12.6,8.5) {\textbf{Meta-prompter} (\method)\\[-1pt]{\scriptsize observes progress and runtimes; names no search technique}};
  \draw[rounded corners=5pt, toolc!60!black, thick, fill=toolc!12] (8.3,-0.3) rectangle (16.9,6.75);
  \node[font=\scriptsize\bfseries, anchor=north west, text=toolc!40!black] at (8.4,6.72) {Coding agent: Claude Code};
  \draw[rounded corners=4pt, black!50, fill=white] (8.5,2.8) rectangle (16.7,6.2);
  \node[font=\scriptsize\bfseries, anchor=north west] at (8.6,6.15) {One rung};
  \node[agentbox, text width=3.0cm] (A) at (12.6,5.45) {Agent proposes and\\implements a \textbf{new technique}};
  \node[toolbox, text width=1.8cm] (B) at (15.5,4.1) {Search with\\solvers};
  \node[trustbox, font=\scriptsize, inner sep=2.5pt, anchor=east] (C) at (15.15,2.32) {\textbf{After the climb:} Lean-proved checkers verify \textbf{certificates}};
  \node[humanbox, text width=1.8cm] (D) at (9.7,4.1) {Measure vs.\\the published\\record};
  \node[plainbox, text width=1.5cm, font=\scriptsize] (T) at (12.6,4.1) {Toolbox\\(only grows)};
  \draw[farr] (A) -| (B);
  \draw[farr] ([xshift=-4mm]B.south) -- ++(0,-0.6) -| (D.south);
  
  \draw[farr] (D.north) |- node[flab, pos=0.3, right] {profile,\\dead ends} (A);
  \draw[farr, dashed, black!50] (D) -- (T);
  \draw[farr, dashed, black!50] (T) -- (B);
  \draw[farr, metad] ([xshift=-5mm]D.north) -- ([xshift=-5mm]D.north |- M.south);
  \node[flab, text=metad, anchor=west] at (9.3,7.45) {progress, runtimes};
  \draw[farr, metad] (M.south -| 13.2,0) -- (A.north -| 13.2,0);
  \node[flab, text=metad, anchor=west] at (13.3,7.45) {``next order'' or\\``new technique''};
  \draw[farr, metad, dashed] (M.west) -| (7.35,7.5);
  \node[flab, text=metad, anchor=east] at (7.3,8.05) {climb to\\$n{+}1$};
  \draw[black!40, dashed] (7.65,5.6) -- (8.5,6.2);
  \draw[black!40, dashed] (7.65,5.6) -- (8.5,2.8);
  \draw[rounded corners=4pt, black!50, fill=white] (8.5,-0.15) rectangle (16.7,1.85);
  \node[font=\scriptsize\bfseries, anchor=north west] at (8.6,1.82) {Solvers and checkers};
  \tikzset{tk/.style={fbox, fill=toolc, font=\tiny, text width=1.8cm, minimum height=0.62cm, inner sep=1.5pt},
           tkn/.style={tk, fill=white, dashed, text=black!60}}
  \node[tk] at (9.55,1.0) {\textbf{CaDiCaL}\\SAT, LRAT proofs};
  \node[tk] at (11.6,1.0) {\textbf{genbg}\\canonical augmentation};
  \node[tk] at (13.65,1.0) {\textbf{Z3}\\SMT};
  \node[tk] at (15.7,1.0) {\textbf{Singular}\\Gr\"obner bases};
  \node[tk, fill=trustc] at (9.55,0.25) {\textbf{Lean 4}\\verified checkers};
  \node[tk, fill=trustc] at (11.6,0.25) {\textbf{LRAT checker}\\verified in Lean};
  \node[tk] at (13.65,0.25) {\textbf{labelg}\\isomorphism classes};
  \node[tk, fill=white] at (15.7,0.25) {\textbf{C++ / Python}\\the agent's code};
  \draw[farr, toolc!40!black] (B.south) -- (15.5,1.85) node[flab, pos=0.72, right] {calls};
  \tikzset{lg/.style={draw, minimum width=3mm, minimum height=2.5mm, inner sep=0pt}}
  \node[lg, fill=metac] (l0) at (-1.6,-1.0) {}; \node[font=\scriptsize, anchor=west] at (l0.east) {meta-prompter};
  \node[lg, fill=agentc] (l1) at (2.0,-1.0) {}; \node[font=\scriptsize, anchor=west] at (l1.east) {coding agent};
  \node[lg, fill=toolc] (l2) at (5.6,-1.0) {}; \node[font=\scriptsize, anchor=west] at (l2.east) {search tools};
  \node[lg, fill=trustc] (l3) at (9.8,-1.0) {}; \node[font=\scriptsize, anchor=west] at (l3.east) {Lean / certificate checking};
  \node[lg, fill=humanc] (l4) at (14.2,-1.0) {}; \node[font=\scriptsize, anchor=west] at (l4.east) {human baseline};
\end{tikzpicture}}
\caption{\method\ on the minimum KS problem. \emph{Left:} the agent climbs one order $n$ at a time. At least one new technique clears each rung (bold: classed as new in Section~\ref{sec:analysis}). Below the published bound of 24 the pipeline must reproduce the published record. The left column gives its speed-up over SAT+CAS (Table~\ref{tab:bench}). Certificate checks cover orders 1--24 (green bar). They yield the lower bound of 25 under the trusted base of Section~\ref{sec:cert}. \emph{Right:} the harness. The meta-prompter observes progress and runtimes and sends short prompts: status checks, the next order with a runtime target, and requests for new techniques. The coding agent (Claude Code) proposes and implements a technique, searches with the solvers, and measures the result against the published record. Certification came after the climb and used checkers proved sound in Lean. Profiles and dead ends seed the next rung, and kept techniques accumulate in a toolbox.}
\label{fig:ladder}\label{fig:loop}
\end{figure}

\paragraph{Our approach.} \method\ (Figure~\ref{fig:ladder}, Section~\ref{sec:method}) makes the agent earn each problem size with one new, measured technique. Thus every gain is attributable, and the protocol keeps rejected ideas as results. The agent may write the search, the certificate generators and the checkers. The result depends on Lean proofs of checker soundness and on complete accepted checks, not on the search implementation. Compiled checker execution, file handling and assembly of the verification records remain in the trusted base (Section~\ref{sec:cert}).

\paragraph{Results.} On orders 17--23 the agent's pipeline is $4\times$ to $2.5\times10^{4}\times$ faster than SAT+CAS on one laptop. The gain grows with the order, and the non-colourable candidate counts match the published counts exactly. The principal result is \lean{NoKSUpTo 24}: every KS set in $\R^3$ has at least 25 vectors. It follows from Lean-proved soundness theorems and from accepted enumeration, per-structure and obstruction checks, under the explicit trusted base of Section~\ref{sec:cert}. The retained order-24 records account for every required structure and cube partition. Table~\ref{tab:status} lists the results through order 24 with their runtimes and verification status.

\needspace{5\baselineskip}
\paragraph{Contributions.}
\begin{enumerate}
\item \textbf{\method}, a protocol and harness for agent-driven technique discovery on exact search problems. It has one measured, new technique per rung, a ratchet with recorded dead ends, and a verification gate (Section~\ref{sec:method}).
\item \textbf{A benchmark result}: a pipeline up to $2.5\times10^4$ times faster than the state of the art. Exact agreement with published candidate counts validates it (RQ1, Section~\ref{sec:bench}).
\item \textbf{A certified lower bound of 25}: \lean{NoKSUpTo 24}. We establish it with checkers proved sound in Lean and with accepted checks that cover 2{,}200{,}309 final basis structures over orders 1--24. The order-24 verification record has complete case coverage. Its checks took 46.6 \cpu-hours on one laptop. The trusted base includes compiled execution, file handling and log assembly (RQ2, Sections~\ref{sec:cert} and~\ref{sec:open}).
\item \textbf{An analysis of novelty}. An ablation attributes the speed-up to individual techniques. A prior-art audit classes each technique as known, transferred or new (RQ3--RQ5, Section~\ref{sec:analysis}).
\end{enumerate}

\begin{table}[t]
\caption{Positioning. ``New result'': a previously unknown mathematical result or record. $^{a}$DRAT proofs cover the SAT calls. The computer-algebra and enumeration steps are not proof-checked. $^{b}$The paper describes how DRAT proofs could be produced and estimates about 5\,TB for order 23. The Z3 embeddability steps have no checkable proofs. $^{c}$Orders up to 24: a checker proved sound in Lean checks every step. The checks are compiled executions assembled from logs, not one kernel-checked proof term (Section~\ref{sec:cert}).}
\label{tab:position}
\centering\footnotesize
\setlength{\tabcolsep}{4pt}
\resizebox{\linewidth}{!}{\begin{tabular}{lcccccc}
\toprule
 & AI proposes & non-existence & new & all steps & per-technique & dead ends\\
 & techniques & result & result & machine-checked & attribution & reported\\
\midrule
SAT+CAS~\citep{li2024} & \texttimes & \checkmark & \checkmark & partial$^{a}$ & --- & ---\\
SMS~\citep{kirchweger2023} & \texttimes & \checkmark & \checkmark & \texttimes$^{b}$ & --- & ---\\
FunSearch, AlphaEvolve~\citep{romeraparedes2024,novikov2025} & \checkmark & \texttimes & \checkmark & \texttimes & \texttimes & \texttimes\\
TTT-Discover~\citep{yuksekgonul2026} & \checkmark & \texttimes & \checkmark & \texttimes & partial & \checkmark\\
LLM-LNS~\citep{ye2025llmlns} & \checkmark & \texttimes & \texttimes & \texttimes & \checkmark & \texttimes\\
WZ + LLM + Lean~\citep{xiong2026wz} & partial & \texttimes & \texttimes & \checkmark & \checkmark & \texttimes\\
\midrule
\method\ (ours) & \checkmark & \checkmark & \checkmark & \checkmark$^{c}$ & \checkmark & \checkmark\\
\bottomrule
\end{tabular}}
\end{table}

\section{Background and related work}
\label{sec:bg}

\paragraph{KS sets.} A finite set $V\subset\R^3$ is \emph{admissible} if its vectors are pairwise linearly independent. A \emph{010-colouring} maps $V$ to $\{0,1\}$ so that no two orthogonal vectors are both 1 and every orthogonal basis in $V$ contains exactly one 1. A \emph{KS set} is an admissible set with no 010-colouring. Its \emph{orthogonality graph} has an edge per orthogonal pair, and its triangles are bases. The orthogonality graph of a minimum KS set has four properties~\citep{arends2011,li2024}. It is squarefree. It has minimum degree 3, with every vertex in a triangle. It is not 010-colourable. It \emph{embeds}: some nonzero, pairwise non-parallel vectors realise its edges as orthogonalities. Non-embeddability is the hard, real-algebraic part. Appendix~\ref{app:glossary} collects the terms used in this paper.

\paragraph{Previous searches.} \citet{arends2011} enumerated squarefree graphs with every vertex in a triangle and tested embeddability. \citet{uijlen2016} restricted the class further and improved the embeddability test. \citet{li2024} ran orderly generation inside a SAT solver through IPASIR-UP~\citep{fazekas2023}. They parallelised the search by cube-and-conquer and checked surviving candidates with computer algebra. \citet{kirchweger2023} used SMS~\citep{kirchweger2021} with \emph{co-certificate learning}: a 010-colouring of a candidate becomes a clause that blocks every candidate it colours. \citet{siddique2026} check the non-embeddability half of the bound of 24 in Lean.

\paragraph{Certified search.} CDCL solvers such as CaDiCaL~\citep{biere2024} log refutations as DRAT or LRAT proofs, and small verified checkers validate these proofs~\citep{cruzfilipe2017}. Cube-and-conquer~\citep{heule2011} made this work for the Boolean Pythagorean triples~\citep{heule2016}. For the empty hexagon number, \citet{subercaseaux2024} proved the reduction and the encoding sound in Lean. They checked the SAT proofs with the verified checker \texttt{cake\_lpr}. Lean~4~\citep{demoura2021} can run proved-correct functions as compiled code, which puts the compiler in the trusted base. \citet{xiong2026wz} pair computer-algebra certificates with Lean for known combinatorial identities.

\paragraph{AI for discovery.} FunSearch~\citep{romeraparedes2024} and AlphaEvolve~\citep{novikov2025} evolve programs that construct objects scored by an evaluator. TTT-Discover~\citep{yuksekgonul2026} trains at test time on such a reward. All three need a continuous score for a candidate construction, and TTT-Discover names sparse, binary rewards as open. LLM-guided search also produced solver components, such as neighbourhoods for MILP~\citep{ye2025llmlns}. The AI Scientist~\citep{lu2024} automates the loop from idea to paper. \citet{yang2025formal} list the discovery of new mathematics with measurable progress as the highest milestone for formal mathematical reasoning. In \method\ the output is a sequence of search techniques and a non-existence proof. The evaluator measures \cpu\ cost. Certification requires accepted checks with Lean-proved soundness under a stated trusted base.

\section{\method}
\label{sec:method}

\method\ applies to a search problem with an exact answer, a size parameter $n$, and a published cost that grows steeply with $n$. An AI coding agent works on a fixed machine with a shell and a compiler. A human sets the problem and the rules.

\begin{definition}[rung]
At size $n$ the agent holds a pipeline $P_n$ (a set of kept techniques), a profile $\pi_n$ of $P_n$ at size $n$, and a dead-end log $D_n$. A rung step proposes a technique $t$ motivated by $\pi_n$. It validates $P_n\cup\{t\}$ on sizes where the answer is known and measures its cost $c$ at $n{+}1$. The step is \emph{accepted} if $c$ improves on $P_n$ and, where a published record exists, beats that record substantially. Then $P_{n+1}=P_n\cup\{t\}$. Otherwise the agent appends $(t,c)$ to $D_n$. A result at size $n$ is \emph{certified} only when two conditions hold. First, a proof assistant proves that the checkers are sound. Second, these checkers accept certificates that cover every case, under a stated trusted base (Section~\ref{sec:cert}). Otherwise we report the result as a computational result.
\end{definition}

\noindent In this study the Lean certification of orders 1--24 came after the climb, when the pipeline was fixed. The definition fixes six rules:
\begin{enumerate}
\item \textbf{At least one new technique per rung.} A measurement motivates each technique: a profile, a slow tail of instances, or a failed idea.
\item \textbf{Validate below, measure at the rung.} At smaller orders the answers and intermediate counts are known, so these orders catch errors before the errors cost time.
\item \textbf{Ratchet with dead ends.} Kept techniques stay. The agent records every rejected idea with its measurement.
\item \textbf{Beat the record} substantially, where one exists. In this study the pipeline beat the record only from order 20 on. The first pipeline (orders 17--19) was slower than SAT+CAS. At order 21, the structure-first pipeline replaced the earlier orderly-generation pipeline.
\item \textbf{Honest cost.} Every run reports wall-clock time and total \cpu\ (user+sys) on one fixed machine. Here the machine is an Apple M3 laptop with 8 cores and 16\,GB.
\item \textbf{Verification gate.} The agent may write generators and checkers. A checker counts only when its soundness is a Lean theorem, and no search tool is trusted.
\end{enumerate}
Rules 1 and 5 stop the agent from scaling a fixed method with more compute. For each rung, Rules 1--3 produce a measured bottleneck, the technique that answers it, and its gain. This is the raw material of Section~\ref{sec:analysis}. Rule~6 separates ``is the claim true?'' from ``is the agent's code correct?''.
\paragraph{The harness.} The coding agent is Claude Code (\texttt{claude-opus-5-5}). It calls the solvers and checkers directly from the shell. Certification requires accepted checks by checkers proved sound in Lean. The \emph{meta-prompter} is a second Claude Code agent, also \texttt{claude-opus-5-5}. It read the literature on the KS problem and related combinatorial searches. It watches progress and runtimes and drives the coding agent's session through computer use. Its prompts are deliberately simple. Of its 57 messages, 12 set the next order and a runtime target, 4 asked for new techniques, and 21 were status checks. A further 6 prepared the tools and gave the measurement rules, and 3 answered the agent's questions. The other 11 were more directive: for example, they suggested a search for patterns in the previous order's candidates, or a positive control. The meta-prompter named no search technique (its setup messages did choose CaDiCaL as the solver) and wrote no code. The broader exploratory session ran over ten days. With verification and writing, it used 5.9\,M output tokens and $1.1\times10^9$ cached input tokens. These session totals include exploratory work outside the certified results reported here.

\section{The pipeline the agent built}
\label{sec:ks}

Table~\ref{tab:ladder} lists, for each order, the measurement that motivated the new technique and the cost of the pipeline at that point.
Orders 17--20 used a first pipeline that followed the published approach: orderly generation inside CaDiCaL, colouring clauses learned during search, and cube-and-conquer. It reached a factor of about 2.5 over SAT+CAS at order 20 and no further. Its order-20 profile showed that CaDiCaL's propagation used the time, not the agent's code. Thus a faster implementation of the same search could not help: the \emph{search space} had to change. That diagnosis led to the structure-first search at order 21.

\begin{table}[t]
\caption{The rungs. Bold: classed as new in Table~\ref{tab:novelty}. \cpu\ is user+sys for the whole run on the laptop, with the pipeline as it was when the agent cleared the rung. Order 23 includes discovering two obstructions. $^\dagger$Shortcuts argued on paper, used only in the uncertified search.}
\label{tab:ladder}
\centering\scriptsize
\setlength{\tabcolsep}{3pt}
\begin{tabularx}{\linewidth}{c >{\raggedright\arraybackslash}X >{\raggedright\arraybackslash}X r r}
\toprule
$n$ & measurement at rung $n{-}1$ & new technique at rung $n$ & \cpu\ (s) & gain vs SAT+CAS\\
\midrule
17 & --- (start) & orderly generation in CaDiCaL (IPASIR-UP); colouring CEGAR & 21 & $0.9\times$\\
18 & one solver leaves 7 cores idle & cube-and-conquer on canonical prefixes; shared colouring clauses & 168 & $0.6\times$\\
19 & the same non-embeddable subgraphs are re-found in many cubes & prefix-time pruning with a library of Z3-proven obstructions & 1{,}716 & $0.3\times$\\
20 & 79\% of time in cubes whose vertex 0 lies in $\le 2$ triangles & \textbf{two-basis theorem} (Thm.~\ref{thm:tb}), dominance lemmas, dynamic splitting & 3{,}308 & $2.5\times$\\
\midrule
21 & time is in CaDiCaL propagation; a 1/10 sample projects to $\approx10^5$\,s & \textbf{structure-first search}: enumerate basis structures with \texttt{genbg} and hereditary pruning; one small SAT problem per structure & 582 & $201\times$\\
22 & 98\% of time builds clauses in Python; worst structures are highly symmetric & C++ solver linked to CaDiCaL; residual symmetry breaking under $\mathrm{Aut}(G_B)$; one-pass enumeration & 317 & $1.1{\times}10^4$\\
23 & encoding is $2/3$ of solver time ($\approx$28k clauses per structure) & lazy squarefree constraints; on-the-fly obstruction discovery (numerical shrinking + Z3) & 1{,}767 & $2.5{\times}10^4$\\
\midrule
24 & colourings are learned one SAT call at a time: 5.8\,M calls and 3.6\,M colouring clauses at order 23 & \textbf{exact-transversal clauses}; edge-minimality pair filter$^\dagger$; solver-free precheck$^\dagger$ & 3{,}623 & ---\\
\bottomrule
\end{tabularx}
\end{table}

\paragraph{Basis structures.} Write $G=G_B+X$. Here $G_B$ is the union of the triangles of $G$, recorded as the 3-uniform hypergraph of its bases. $X$ is the set of remaining edges. $G_B$ is linear (two bases share at most one vector), its collinearity graph is squarefree, and its only triangles are bases. We call such a hypergraph a \emph{basis structure}. $X$ must add no triangle and no 4-cycle.

\paragraph{Stage 1: enumerate structures.} nauty's bipartite generator \texttt{genbg}~\citep{mckay2014} generates vector--basis incidence graphs up to isomorphism by canonical augmentation~\citep{mckay1998}. The agent added hooks that reject a partial structure as soon as it has a 4-cycle or a non-basis triangle. Both conditions are hereditary, so the enumeration loses no structure.

\paragraph{Stage 2: one SAT problem per structure.} CaDiCaL searches over one variable per non-basis vertex pair. The clauses require no new triangle, no new 4-cycle and minimum degree~3. The solver also learns three kinds of constraint.
\emph{Colourings}: a 010-colouring with ones on $S$ yields ``some edge of $X$ lies inside $S$''. These are co-certificates in the sense of \citet{kirchweger2023}. From order 24 on, the stage-2 program adds them up front as \emph{exact transversals} of $G_B$ (one vector from each basis).
\emph{Obstructions}: the stage-2 program matches a non-colourable model against a library of non-embeddable graphs, and a match forbids the matched edges.
\emph{Lazy squarefree constraints}: the stage-2 program adds a clause with more than two free literals only when a model violates that clause.
When the structure is fixed, every constraint is invariant under $\mathrm{Aut}(G_B)$. Lex-leader constraints~\citep{crawford1996} for its generators therefore break the symmetry.

The order-20 technique is a computer-assisted theorem on KS sets in which every vector lies in at most two bases (Theorem~\ref{thm:tb}, Appendix~\ref{app:tb}, not certified in Lean). The structure-first search later subsumed it.

\begin{figure}[t]
\centering
\resizebox{\linewidth}{!}{
\begin{tikzpicture}[x=1cm,y=1cm]
  \definecolor{casc}{HTML}{F3E1A6}
  \tikzset{pb/.style={fbox, text width=#1, font=\scriptsize}, lb/.style={pb=#1, fill=trustc, minimum height=1.05cm}, pb/.default=2.6cm,
           fam/.style={font=\scriptsize\bfseries, anchor=south, text=black!65},
           larr/.style={-{Stealth[length=1.6mm]}, thick, trustd, dashed}}
  \node[fam] at (1.4,2.6) {1. enumerate};
  \node[pb=2.6cm, fill=toolc] (E) at (1.4,1.6) {\textbf{nauty \texttt{genbg}}: canonical augmentation, hereditary pruning\\[1pt]\textbf{1{,}748{,}732} basis structures};
  \node[fam] at (5.3,2.6) {2. SAT per structure};
  \node[pb=3.0cm, fill=toolc] (S) at (5.3,1.6) {\textbf{CaDiCaL}: no new triangle or 4-cycle, degree $\ge3$; lex-leader under $\mathrm{Aut}(G_B)$; exact transversals; lazy squarefree\\[1pt]\textbf{8.2\,M} SAT calls};
  \node[fam] at (10.5,2.6) {3. obstruction library};
  \node[pb=2.6cm, fill=toolc] (L) at (10.5,1.6) {match a known non-embeddable subgraph; forbid its edges and re-solve\\[1pt]closes \textbf{89{,}593} (5.1\%)};
  \node[fam] at (14.5,2.6) {4. computer algebra};
  \node[pb=2.6cm, fill=casc] (C) at (14.5,1.6) {numeric shrink, then \textbf{Z3} \texttt{nlsat} or \textbf{Singular} Gr\"obner bases\\[1pt]\textbf{0} calls at order 24};
  \node[pb=4.2cm, fill=white] (U) at (8.0,0.1) {no non-colourable completion: \textbf{1{,}292{,}722} by SAT (73.9\%), \textbf{366{,}417} by precheck$^\dagger$ (21.0\%)};
  \draw[farr] (E) -- node[flab, above] {each} (S);
  \draw[farr] (S.east) -- node[flab, above, pos=0.5] {non-colourable\\model} (L.west);
  \draw[farr] (U.north |- S.east) -- node[flab, right] {UNSAT} (U.north);
  \draw[farr] (L.east) -- node[flab, above] {no match} (C.west);
  \draw[farr] (C.south) -- ++(0,-0.35) -| node[flab, pos=0.25, below] {new obstruction} ([xshift=8mm]L.south);
  \draw[rounded corners=4pt, trustd, fill=trustc!35] (-0.05,-2.6) rectangle (16.0,-0.5);
  \node[font=\scriptsize\bfseries, anchor=south west, text=trustd] at (0.0,-2.58) {Lean 4 (trusted base: Section~\ref{sec:cert})};
  \node[lb=2.6cm] (LE) at (1.4,-1.2) {\lean{enumCheckF}: the enumeration is complete};
  \node[lb=3.0cm] (LS) at (5.3,-1.2) {\lean{checkSym}: 1{,}748{,}668 structures by LRAT; \lean{checkSymSplit}: 65 in 45{,}870 cubes};
  \node[lb=2.6cm] (LO) at (10.5,-1.2) {\lean{obsCheck}: every graph in $L^{\ast\ast}$ is non-embeddable};
  \node[pb=2.6cm, fill=white] (TH) at (14.5,-1.2) {\lean{NoKSUpTo 24}\\(Theorem~\ref{thm:main})};
  \draw[larr] (E.south) -- node[flab, right, text=trustd] {certificate} (LE.north);
  \draw[larr] (S.south) -- node[flab, left, text=trustd] {LRAT proofs} (LS.north);
  \draw[larr] (L.south) -- node[flab, right, text=trustd] {certificates} (LO.north);
  \draw[thick, trustd] (LE.south) -- ++(0,-0.22) coordinate (b1);
  \draw[thick, trustd] (LS.south) -- (LS.south |- b1);
  \draw[thick, trustd] (LO.south) -- (LO.south |- b1);
  \draw[farr, trustd] (b1) -| (TH.south);
\end{tikzpicture}}
\caption{The pipeline that solved order 24. Top: the search run (923\,s wall, 3{,}623\,s \cpu). Computer algebra: Z3's \texttt{nlsat}~\citep{demoura2008,jovanovic2012} or Singular~\citep{singular}. Bottom: the certified run, without shortcuts (it also keeps one 8-basis structure that the search skips).}
\label{fig:pipe24}
\end{figure}
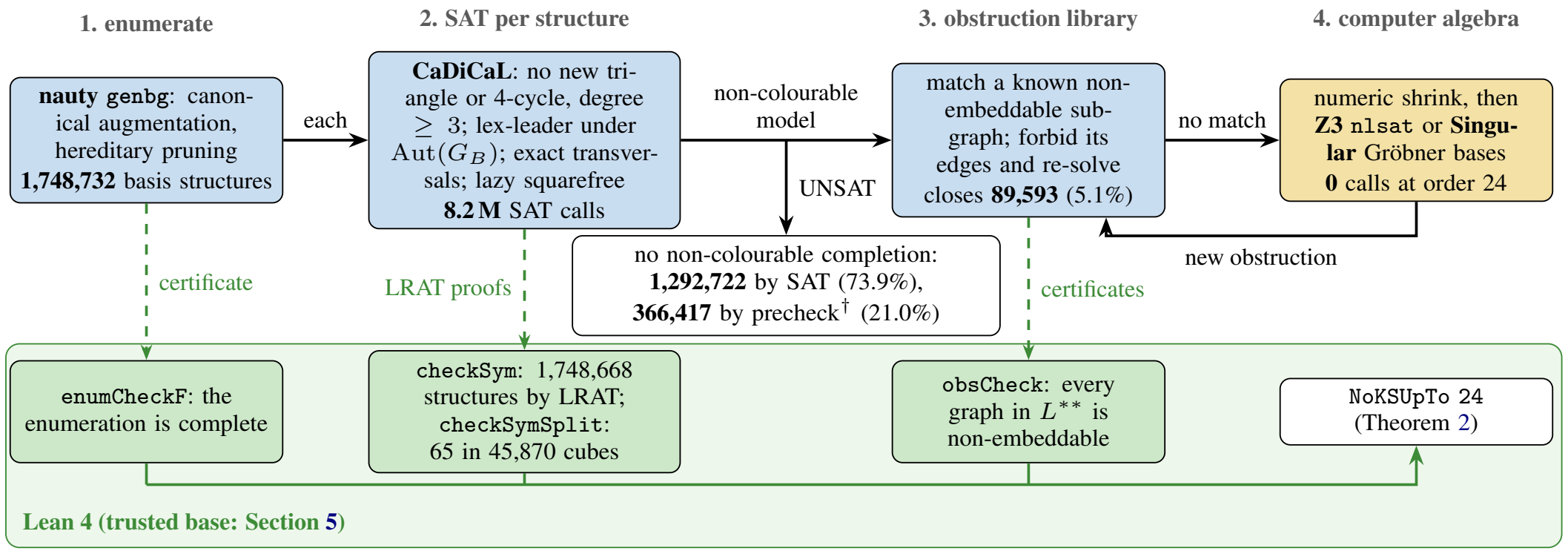

\paragraph{Which solver did the work?} Orderly generation inside the SAT solver, the design of SAT+CAS and SMS, first solved orders 17--20. From order 21 on, nauty enumerates the structures and CaDiCaL solves one small problem per structure. The pipeline calls computer algebra only for a non-colourable graph that no known obstruction explains. At order 24 (Figure~\ref{fig:pipe24}) SAT closes 73.9\% of the structures and a solver-free precheck$^\dagger$ closes 21.0\%. The obstruction library (built by Z3 at earlier rungs) closes the remaining 5.1\%, and the pipeline never calls computer algebra. Appendix~\ref{app:pipelines} covers every order.

\section{Certification in Lean}
\label{sec:cert}

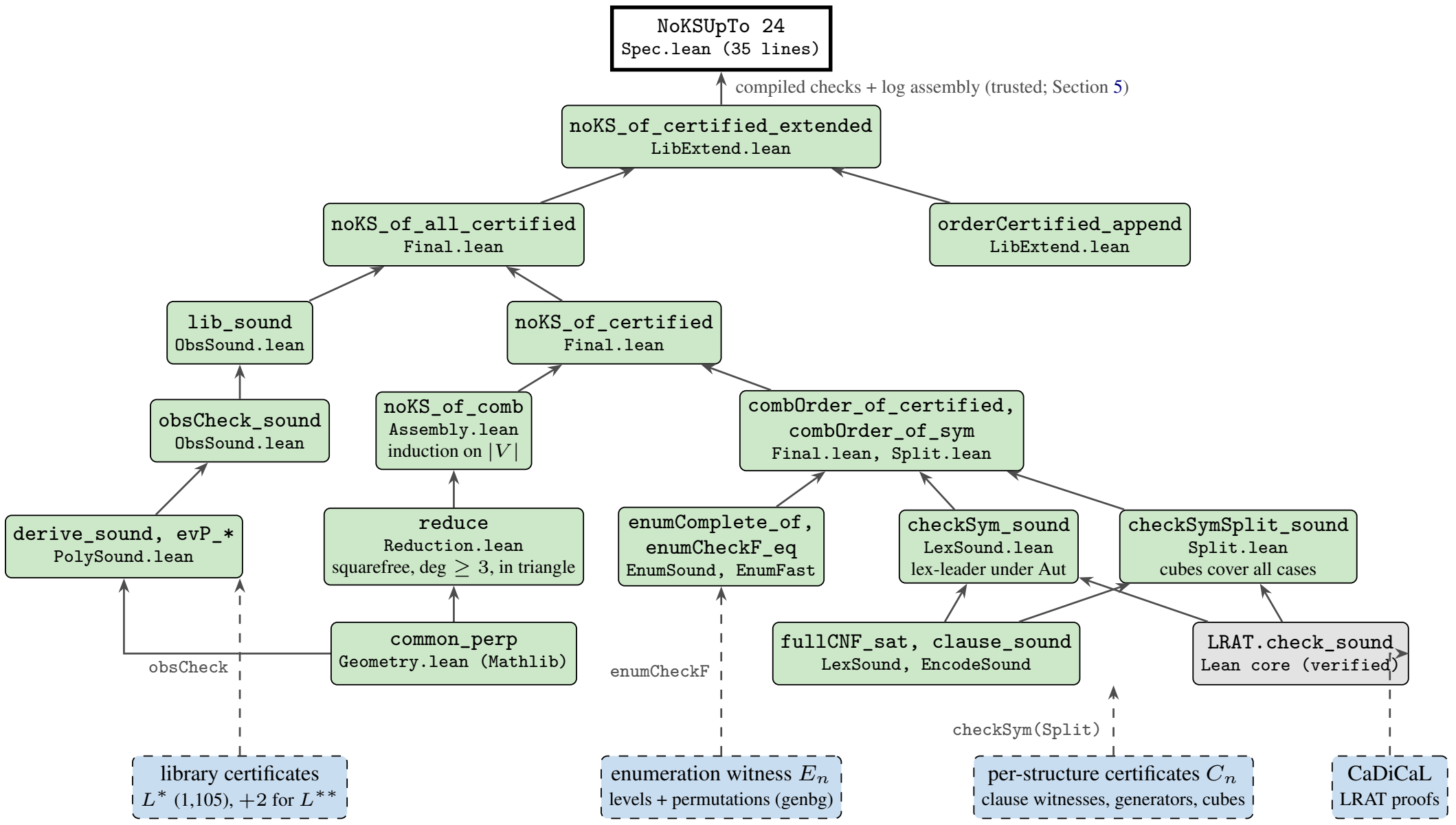
\begin{figure}[t]
\centering
\resizebox{\linewidth}{!}{
\begin{tikzpicture}[x=1cm,y=1cm,
  thm/.style={draw, rounded corners=2pt, fill=trustc, align=center, font=\scriptsize, inner sep=2.5pt, minimum height=7mm},
  stm/.style={draw, very thick, fill=white, align=center, font=\scriptsize, inner sep=3pt, minimum height=7mm},
  crt/.style={draw, dashed, rounded corners=2pt, fill=toolc, align=center, font=\scriptsize, inner sep=2.5pt, minimum height=7mm},
  core/.style={draw, rounded corners=2pt, fill=humanc, align=center, font=\scriptsize, inner sep=2.5pt, minimum height=7mm},
  u/.style={-{Stealth[length=1.6mm]}, semithick, black!70}]
\newcommand{\fl}[1]{\\[-1pt]{\tiny\ttfamily #1}}
\node[stm]  (stmt) at (6.6,9.3) {\ttfamily NoKSUpTo 24\fl{Spec.lean (35 lines)}};
\node[thm]  (ext)  at (6.6,8.2) {\ttfamily noKS\_of\_certified\_extended\fl{LibExtend.lean}};
\node[thm]  (all)  at (3.6,7.1) {\ttfamily noKS\_of\_all\_certified\fl{Final.lean}};
\node[thm]  (app)  at (10.4,7.1) {\ttfamily orderCertified\_append\fl{LibExtend.lean}};
\node[thm]  (lib)  at (1.2,6.0) {\ttfamily lib\_sound\fl{ObsSound.lean}};
\node[thm]  (cert) at (5.4,6.0) {\ttfamily noKS\_of\_certified\fl{Final.lean}};
\node[thm]  (comb) at (3.6,4.9) {\ttfamily noKS\_of\_comb\fl{Assembly.lean}\\[-1pt]\tiny induction on $|V|$};
\node[thm]  (ord)  at (8.4,4.9) {\ttfamily combOrder\_of\_certified,\\ \ttfamily combOrder\_of\_sym\fl{Final.lean, Split.lean}};
\node[thm]  (obs)  at (1.2,4.9) {\ttfamily obsCheck\_sound\fl{ObsSound.lean}};
\node[thm]  (poly) at (-0.1,3.6) {\ttfamily derive\_sound, evP\_*\fl{PolySound.lean}};
\node[thm]  (red)  at (3.6,3.6) {\ttfamily reduce\fl{Reduction.lean}\\[-1pt]\tiny squarefree, deg${}\ge3$, in triangle};
\node[thm]  (geo)  at (3.6,2.4) {\ttfamily common\_perp\fl{Geometry.lean (Mathlib)}};
\node[thm]  (enum) at (6.6,3.6) {\ttfamily enumComplete\_of,\\ \ttfamily enumCheckF\_eq\fl{EnumSound, EnumFast}};
\node[thm]  (sym)  at (9.6,3.6) {\ttfamily checkSym\_sound\fl{LexSound.lean}\\[-1pt]\tiny lex-leader under Aut};
\node[thm]  (spl)  at (12.4,3.6) {\ttfamily checkSymSplit\_sound\fl{Split.lean}\\[-1pt]\tiny cubes cover all cases};
\node[thm]  (cnf)  at (8.9,2.4) {\ttfamily fullCNF\_sat, clause\_sound\fl{LexSound, EncodeSound}};
\node[core] (lrat) at (13.1,2.4) {\ttfamily LRAT.check\_sound\fl{Lean core (verified)}};
\node[crt]  (cl)   at (1.2,0.9) {library certificates\\ \tiny $L^\ast$ (1{,}105), $+2$ for $L^{\ast\ast}$};
\node[crt]  (ce)   at (6.6,0.9) {enumeration witness $E_n$\\ \tiny levels + permutations (genbg)};
\node[crt]  (cs)   at (11.0,0.9) {per-structure certificates $C_n$\\ \tiny clause witnesses, generators, cubes};
\node[crt]  (cd)   at (14.1,0.9) {CaDiCaL\\ \tiny LRAT proofs};
\draw[u] (ext) -- node[right, font=\tiny, xshift=1pt] {compiled checks + log assembly (trusted; Section~\ref{sec:cert})} (stmt);
\draw[u] (all) -- (ext); \draw[u] (app) -- (ext);
\draw[u] (lib) -- (all); \draw[u] (cert) -- (all);
\draw[u] (comb) -- (cert); \draw[u] (ord) -- (cert);
\draw[u] (obs) -- (lib); \draw[u] (poly) -- (obs); \draw[u] (geo.west) -| (poly.south);
\draw[u] (red) -- (comb); \draw[u] (geo) -- (red);
\draw[u] (enum) -- (ord); \draw[u] (sym) -- (ord); \draw[u] (spl) -- (ord);
\draw[u] (cnf) -- (sym); \draw[u] (cnf) -- (spl); \draw[u] (lrat) -- (sym); \draw[u] (lrat) -- (spl);
\draw[u, dashed] (cl) -- node[left, font=\tiny]{\ttfamily obsCheck} (poly.south -| cl.north);
\draw[u, dashed] (ce) -- node[left, font=\tiny]{\ttfamily enumCheckF} (enum);
\draw[u, dashed] (cs.north) -- node[left, font=\tiny, pos=.35]{\ttfamily checkSym(Split)} (cnf.south -| cs.north);
\draw[u, dashed] (cd.north) |- (lrat.east);
\end{tikzpicture}}
\caption{Blueprint of the certified bound. Arrows point from a result to the results that use it. Each node names its Lean declarations and file under \texttt{lean/SkyLadder/}. Green: proved in Lean. Grey: Lean's own verified LRAT checker. Blue, dashed: certificates from untrusted tools (nauty's \texttt{genbg}, our solver, CaDiCaL, Singular, Z3), accepted only through the named checker. Section~\ref{sec:cert} lists the trusted base.}
\label{fig:blueprint}
\end{figure}

The certification argument separates general soundness theorems from instance-specific checks. Lean proves the reductions and checker soundness. Compiled executions of those checkers accept the enumeration, per-structure and obstruction certificates. The search tools (nauty, the C++ solver, CaDiCaL, Z3, Singular, an SDP solver and Python scripts) only propose certificates. Figure~\ref{fig:blueprint} shows the soundness development and the data that it checks.

\paragraph{Statement.} \lean{NoKSUpTo N} says that no admissible set of at most $N$ vectors in \lean{EuclideanSpace ℝ (Fin 3)} lacks a 010-colouring. It is 35 lines on top of Mathlib~\citep{mathlib2020} (Appendix~\ref{app:spec}). Lean proves the reductions from vectors to basis structures. The certified pipeline uses none of the $\dagger$ shortcuts of Table~\ref{tab:ladder}.

\paragraph{Checkers.} Three checkers carry the computation (details in Appendix~\ref{app:checkers}).
\lean{enumCheckF} accepts a witness that the structure list is complete up to isomorphism.
\lean{checkSym} rebuilds a structure's CNF in Lean and checks the witness of every learned clause and every symmetry generator. It then runs Lean's verified LRAT checker on CaDiCaL's proof (\lean{checkSymSplit} does the same for a cube split).
\lean{obsCheck} accepts Positivstellensatz-style polynomial certificates of non-embeddability.
The conditional soundness theorem, \lean{noKS\_of\_certified\_extended}, derives \lean{NoKSUpTo N} from accepted checks. The soundness development has 21 files and about 3{,}200 lines. It contains no \lean{sorry} and uses only the axioms \lean{propext}, \lean{Classical.choice} and \lean{Quot.sound}. These axiom dependencies describe the soundness proofs. They do not remove the execution assumptions needed to discharge the computational hypotheses of those proofs.

\paragraph{Trusted base.} The result trusts the Lean kernel and Mathlib, the 35-line mathematical specification, and the compiler and runtime that execute the checkers. It also trusts certificate-file handling and the assembly of per-process verification records. For structure checks, process $p$ of $P$ handles indices $\equiv p\pmod P$. Enumeration and cube checks use their documented partitions. The records must accurately describe executions on the stated inputs. We assemble the numerical conclusion from these accepted checks and the soundness theorems. The repository does not contain a single kernel-checked proof term with all data-specific hypotheses of \lean{NoKSUpTo 24} discharged. We do not assume that the search and the certificate generators are correct. The Lean sources, certificate inputs and verification logs are public (see Code availability).

\section{Experiments}
\label{sec:results}
\label{sec:analysis}

We ask five questions.
\textbf{RQ1:} Does the agent's pipeline beat the state of the art on the published orders?
\textbf{RQ2:} Does it resolve the open case with complete certificate coverage and a stated trusted base?
\textbf{RQ3:} Where does the speed come from?
\textbf{RQ4:} Which techniques are new?
\textbf{RQ5:} Did the verification gate change the work?

\paragraph{Setup.} All runs used one Apple M3 laptop (4 performance and 4 efficiency cores, 16\,GB). The software was CaDiCaL 3.0.1, nauty 2.8.8, Z3 4.16, Singular 4.4.1 and Lean~4.28 with Mathlib. One exception: a 4-core cloud machine built the cube tree of one order-24 structure. The baselines are the published runs of SAT+CAS~\citep{li2024} and SMS~\citep{kirchweger2023}. Their orders 17--21 ran sequentially on a 2.1\,GHz Intel Xeon. Their orders 22--23 ran as parallel cube-and-conquer, reported as total \cpu. The SAT+CAS figures include DRAT logging. An M3 core is plausibly 1.5--2.5$\times$ faster per thread, so we report \cpu\ and treat factors below about 3 as ties.

\subsection{\texorpdfstring{RQ1: the agent's pipeline is up to $2.5\times10^4$ times faster, and exact}{RQ1: up to 25,000 times faster, and exact}}
\label{sec:bench}

\begin{figure}[t]
\centering
\includegraphics[width=.92\linewidth]{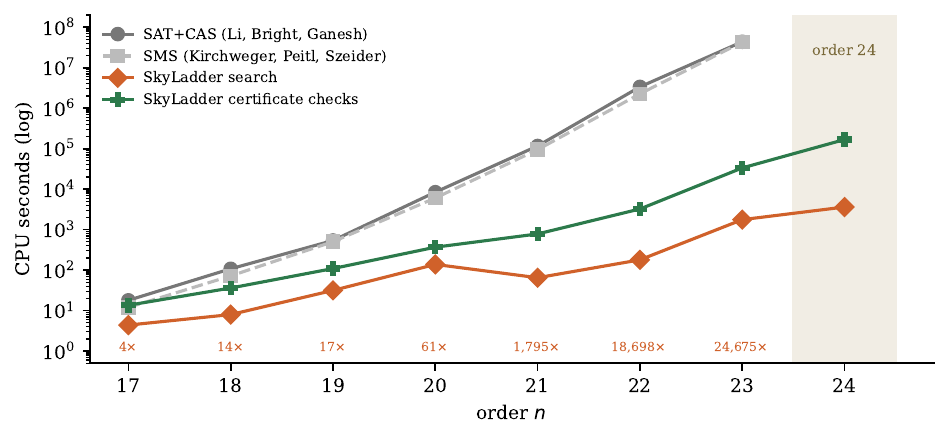}
\caption{Total \cpu\ through order 24 (log scale). Orange: the \method\ search. Green: user+sys \cpu\ of the enumeration and per-structure certificate checks, under the trusted base of Section~\ref{sec:cert}. Grey: SAT+CAS and SMS. Shaded: order 24, which establishes the new lower bound of 25. The numbers along the bottom give the gain of the search over SAT+CAS.}
\label{fig:speed}
\end{figure}

\begin{table}[t]
\caption{Benchmark. Search rows: orders 17--20 use the structure-first pipeline, order 21 the order-22 pipeline, and orders 22--23 the order-23 pipeline (order 23 includes discovering two obstructions). Candidates: isomorphism classes of non-colourable graphs that meet the degree, triangle and squarefree conditions, with embeddability off. Last row: user+sys \cpu\ of Lean's checks. This includes the CaDiCaL runs that produce the LRAT proofs but not certificate generation. Figure~\ref{fig:speed} plots these data.}
\label{tab:bench}
\centering\footnotesize
\setlength{\tabcolsep}{4pt}
\begin{tabular}{lrrrrrrr}
\toprule
order $n$ & 17 & 18 & 19 & 20 & 21 & 22 & 23\\
\midrule
SAT+CAS~\citep{li2024}, \cpu\ s & 18 & 108 & 540 & 8{,}430 & 116{,}700 & $3.36\!\times\!10^6$ & $4.36\!\times\!10^7$\\
SMS~\citep{kirchweger2023}, \cpu\ s & 12 & 72 & 504 & 6{,}048 & 94{,}464 & $2.26\!\times\!10^6$ & $4.29\!\times\!10^7$\\
\method\ search, \cpu\ s & 4.4 & 8.0 & 31.6 & 138.5 & 65.0 & 179.7 & 1{,}767\\
gain vs SAT+CAS & $4\times$ & $14\times$ & $17\times$ & $61\times$ & $1.8{\times}10^3$ & $1.9{\times}10^4$ & $2.5{\times}10^4$\\
\midrule
non-colourable candidates, SMS & 1 & 0 & 8 & 147 & 2{,}497 & 88{,}282 & 3{,}747{,}950\\
non-colourable candidates, ours & 1 & 0 & 8 & 147 & 2{,}497 & 88{,}282 & 3{,}747{,}950\\
\midrule
Lean checks, \cpu\ s & 14 & 36 & 110 & 369 & 784 & 3{,}242 & 33{,}478\\
\bottomrule
\end{tabular}
\end{table}

Table~\ref{tab:bench} gives the comparison.
\textbf{The gain grows with $n$}, from $4\times$ at order 17 to $2.5\times10^4$ at order 23. Orders 17--20 use an earlier, slower version of the pipeline, so their gains are understated. The number of basis structures grows about $2$--$5\times$ per order, while the published cost grows $5$--$30\times$.
\textbf{Candidate counts provide an independent cross-check.} With embeddability switched off, the pipeline enumerates non-colourable candidates. After canonical labelling the counts equal the published SMS counts at every order from 17 to 23. This agreement supports the implementation. The enumeration and per-structure certificate checks establish that the certified result is complete.
\textbf{Certification is cheap.} At order 23 Lean's checks cost $3.3\times10^4$ \cpu-seconds, about 1{,}300 times less than either published search.

\subsection{\texorpdfstring{RQ2: every KS set in $\R^3$ has at least 25 vectors}{RQ2: every KS set in R3 has at least 25 vectors}}
\label{sec:open}

\begin{theorem}[Certified lower bound]\label{thm:main}
Under the trusted base of Section~\ref{sec:cert}, \lean{NoKSUpTo 24}: every Kochen--Specker set in $\R^3$ has at least 25 vectors.
\end{theorem}

The checkers check orders 1--23 against a library $L^\ast$ of 1{,}105 obstructions and order 24 against $L^{\ast\ast}$. $L^{\ast\ast}$ adds two new obstructions (15 and 16 vertices). The agent found them when two structures relied on a library graph that had no certificate. Every library entry has an accepted certificate (1{,}027 real-algebraic, and 80 unused slots that hold $K_4$). A library-extension lemma lifts orders 1--23 to $L^{\ast\ast}$ without a second run.
At order 24, Lean accepted the enumeration (1{,}748{,}733 final structures) and every per-structure certificate, with no failure. The checks took 167{,}884 \cpu-seconds (46.6\,h) on the laptop. Per-structure LRAT proofs closed all but 67 structures. The 67 needed new engineering, and each step was again a measured failure and a targeted fix. The cube-generation scripts partitioned 64 hard structures into 7{,}490 cubes. They partitioned one structure of 8 disjoint bases, with a large symmetry group and little propagation, into 38{,}380 cubes. Two structures needed the new obstructions. Appendix~\ref{app:o24} gives the full accounting and Appendix~\ref{app:blueprint} the certificate data for every order.

\paragraph{Verification coverage.} The 35 primary structure-check parts account for 1{,}748{,}729 distinct order-24 structures. Four supplementary structure identities fill exactly their complement in the 1{,}748{,}733-entry final list, with no overlap or remaining gap. All 12 parts of the 38{,}380-cube check are present, including the coverage check in part 0. The enumeration checks and all 1{,}107 obstruction-library entries also have accepted records. This accounting establishes complete recorded verification coverage. It relies on the execution and provenance assumptions in Section~\ref{sec:cert}.

\begin{table}[t]
\caption{Results through order 24 on the 8-core laptop (\cpu: user+sys). $^a$Certification uses checkers proved sound in Lean, under the trusted base of Section~\ref{sec:cert}. Search times for orders 17--23 use the pipelines of Table~\ref{tab:bench}. The order-24 search time is for the original search with the $\dagger$ shortcuts of Table~\ref{tab:ladder}. The certified order-24 run uses none of those shortcuts. Check \cpu\ covers enumeration and per-structure verification (Table~\ref{tab:perorder} lists user \cpu\ only).}
\label{tab:status}
\centering\footnotesize
\setlength{\tabcolsep}{4pt}
\resizebox{\linewidth}{!}{\begin{tabular}{lllll}
\toprule
order $n$ & exclusion result & search: wall / \cpu & verification & check \cpu\\
\midrule
1--16 & no KS set & --- & \textbf{certified}$^a$ & 9\,s\\
17--23 & no KS set; candidate counts $=$ SMS & Table~\ref{tab:bench} & \textbf{certified}$^a$ & 10.6\,h\\
24 & no KS set (\emph{new}) & 15\,min / 1.0\,h & \textbf{certified}$^a$ (Thm.~\ref{thm:main}) & 46.6\,h\\
\bottomrule
\end{tabular}}
\end{table}

\subsection{RQ3: one reformulation, then engineering}

Figure~\ref{fig:ablation} (Appendix~\ref{app:dead}) shows the measured effect of each kept technique on the same instance before and after. We measured the rows on different orders, so they are not additive.
The structure-first search alone gives $24\times$ at order 20 over the best orderly-generation pipeline. At order 21 it gives about $170\times$ (projected $10^5$ to 582 \cpu-seconds). Every other technique gives $1.8$--$9.5\times$.
The large gain therefore comes from a change in \emph{what is searched}. Both baselines use orderly generation, which builds graphs vertex by vertex. It must rediscover the basis structure of each candidate many times inside the solver. Our pipeline enumerates basis structures first, with a mature isomorph-free generator. This removes that redundancy and leaves a small problem per structure (at order 24, 8.22 million SAT calls for 1.75 million structures).
After the reformulation, most search work scales with the number of structures. A thin tail of sparse, highly symmetric structures dominates the difficult cases. At order 24, certification of this tail requires the cube splits and supplementary checks described in Appendix~\ref{app:o24}.

\subsection{RQ4: a few new ideas, several transferred, some rediscovered}

\begin{table}[t]
\caption{Novelty audit of the kept techniques of rungs 17--24. ``cert.'': part of the certified pipeline only.}
\label{tab:novelty}
\centering\scriptsize
\setlength{\tabcolsep}{3pt}\renewcommand{\arraystretch}{1.12}
\begin{tabularx}{\linewidth}{>{\raggedright\arraybackslash}p{30mm} c >{\raggedright\arraybackslash}X}
\toprule
technique (rung) & class & closest prior work and what differs\\
\midrule
orderly generation in the solver (17) & known & \citet{li2024}; SMS~\citep{kirchweger2021}\\
colouring CEGAR (17) & known & co-certificate learning~\citep{kirchweger2023}\\
cube-and-conquer (18) & known & \citet{heule2011}; used by both baselines\\
obstruction library as learned clauses (19, 23) & transferred & to our knowledge the baselines check embeddability after the search; here obstructions are clauses inside the search, found and proved on the fly\\
two-basis theorem (20) & new & a new computer-assisted result (Thm.~\ref{thm:tb})\\
\textbf{structure-first search} (21) & new & earlier work restricts to graphs with every vertex in a triangle~\citep{arends2011,uijlen2016} but enumerates graphs; we enumerate the basis hypergraph with \texttt{genbg} and hereditary pruning, then solve one SAT problem per structure\\
residual symmetry breaking (22) & transferred & lex-leader~\citep{crawford1996} and canonical augmentation~\citep{mckay1998} are standard; combining them (isomorph-free on the structure, lex-leader on the completion) is new here\\
lazy squarefree constraints (23) & transferred & lazy clause generation; \citet{li2024} encode squarefreeness with static clauses\\
exact-transversal clauses (24) & new & SMS learns colourings one at a time; here every colouring of the structure is enumerated as an exact transversal of $G_B$ and added up front\\
polynomial obstruction certificates in Lean (cert.) & transferred & concurrent with \citet{siddique2026}; computer-algebra embeddability checks are used by \citet{li2024}\\
certified isomorph-free enumeration (cert.) & new & to our knowledge no proof-assistant check of a canonical-augmentation enumeration at this scale; \citet{subercaseaux2024} needed no isomorph rejection\\
library extension; cube-index splitting (24, cert.) & engineering & small, but necessary for the result\\
\bottomrule
\end{tabularx}
\end{table}

Table~\ref{tab:novelty} classes each kept technique against the closest prior work that we know. The classes are \emph{known} (already used for this problem), \emph{transferred} (standard elsewhere but, to our knowledge, new here) and \emph{new} (no precedent found). The audit is ours, not the agent's. We did not use independent raters, and we invite corrections, especially against \citet{arends2011} and \citet{uijlen2016}.
The picture is mixed. Most rungs rediscovered or transferred known ideas. Rungs 17--19 rebuilt the published approach and ran slower than it. The decisive idea, the structure-first search, is new for this problem as far as we can tell. However, it combines a known reduction with a known generator and is not an exotic invention. The protocol added the \emph{diagnosis} that led to it.
At rungs 17--24 the agent also recorded 25 rejected ideas with their measurements, about as many as it kept (Appendix~\ref{app:dead}). Examples are an auxiliary-variable squarefree encoding ($+9\%$), the full obstruction library instead of its 40 most useful entries ($+29\%$), and fully lazy squarefree constraints (944 million lazy clauses).

\subsection{RQ5: verification changed the work, not only the claim}
Lean rejected certificates twice during development. Both times the cause was a real bug in the agent's generator. The first bug was self-pair literals in transversal and degree clauses. The second was 4-cycle clauses that the generator used but did not print. Both bugs were in the certificate generator, not the search, and made certificates incomplete, not wrong. The candidate-count check tests the search, so it could not find these bugs. Lean rejected all five deliberately tampered obstruction certificates. Verified symmetry breaking also made certification feasible. Without it, an order-18 structure of 6 disjoint bases ran for more than 42 minutes and produced a 977\,MB proof. With it, the structure closes at once.

\section{Limitations}
\label{sec:limits}

Table~\ref{tab:evidence} grades the evidence behind each claim.

\begin{table}[ht]
\caption{Strength of evidence for each claim.}
\label{tab:evidence}
\centering\footnotesize
\setlength{\tabcolsep}{4pt}
\begin{tabularx}{\linewidth}{>{\raggedright\arraybackslash}p{52mm} >{\raggedright\arraybackslash}X}
\toprule
claim & evidence\\
\midrule
no KS set with $\le 24$ vectors & accepted checks with Lean-proved soundness and complete recorded coverage, under the trusted base of Section~\ref{sec:cert}\\
candidate-count reproduction, orders 17--23 & exact agreement with independently published counts; a cross-check, not the certification argument\\
speed-ups over SAT+CAS and SMS & single timed runs, compared with published runs on other hardware\\
novelty classes (Table~\ref{tab:novelty}) & the authors' literature audit; no independent raters\\
\method\ as a protocol & one exploratory climb on one problem, not preregistered\\
\bottomrule
\end{tabularx}
\end{table}

\begin{itemize}
\item \textbf{Trusted base.} Certification relies on compiled checker execution, file handling and accurate assembly of the retained records. It also relies on the Lean kernel, Mathlib and the mathematical specification. The numerical bound is not packaged as one fully instantiated kernel-checked proof term. Complete log accounting establishes coverage, not independent replay of every historical proof check.
\item \textbf{Hardware.} The gaps at orders 18--23 exceed any plausible hardware factor. The $4\times$ at order 17 does not.
\item \textbf{Attribution.} The meta-prompter named no search technique, so we attribute the techniques to the coding agent. A human chose the problem and set the rules. Some messages did set a direction. For example, the two-basis theorem followed a suggestion to search for patterns and prove a conjecture, though the statement and proof are the agent's. We did not measure how the result depends on the meta-prompter's wording or on the choice of coding agent.
\item \textbf{Generality.} We applied \method\ to one problem. Transfer to other searches is open.
\end{itemize}

\section{Conclusion}
\method\ establishes that every Kochen--Specker set in $\R^3$ has at least 25 vectors. It uses certificate checkers proved sound in Lean and complete recorded verification coverage through order 24. The trusted base includes compiled checker execution, file handling and assembly of the accepted checks. The structure-first reformulation also reproduces the published candidate counts. It yields a search up to $2.5\times10^4$ times faster at the last published order. The central contribution combines measured technique discovery with a certification argument that covers enumeration, SAT refutations and real-algebraic obstructions. A fully instantiated proof term for the accepted computations would further reduce reliance on external assembly.

\bibliographystyle{plainnat}
\bibliography{refs}

\section*{Code availability} The proof artifacts for orders 1--24 are public at \url{https://github.com/BrianLi009/KS24} (release \texttt{v1.0.0}, commit \texttt{88c9702}). The repository contains four groups of files. \texttt{lean/} has the Lean soundness development and the source of the compiled checker. \texttt{data/} has both obstruction libraries and the certificate blocks for the 67 hard cases of order 24. \texttt{logs/} has the retained verification logs for every order. \texttt{audit/coverage.json} is a coverage audit that reconciles those logs with the final structure lists by exact identity. The large certificate inputs (98 files, 11.0\,GB decoded) are release assets, each with a SHA-256 hash in the repository's manifest. These hashes bind the published files. The historical logs do not record a hash of every input for every run. The checking driver removed the temporary LRAT files after it checked them, and we did not archive these proofs. A new run of the checkers regenerates and verifies them, and the repository gives the full 103-job recheck plan. We did not repeat that full recheck for the release. The search code and certificate generators are available from the authors on request. We will release them publicly in a later version of this paper.

\section*{Use of AI} The AI agent described in Section~\ref{sec:method} wrote the search code, the certificate generators and most of the Lean development. This is the subject of the paper.

\appendix
\section{Glossary}
\label{app:glossary}
\begin{description}[leftmargin=0pt,itemsep=1pt]
\item[010-colouring.] A map $f:V\to\{0,1\}$ with no two orthogonal vectors both 1 and exactly one 1 in every orthogonal basis.
\item[KS set.] An admissible (pairwise non-parallel) set of vectors with no 010-colouring.
\item[Order $n$.] The number of vectors. ``Order $n$ is certified'' means that checkers proved sound in Lean accepted complete certificates. These certificates exclude every KS set with exactly $n$ vectors, under the trusted base of Section~\ref{sec:cert}.
\item[Embedding.] Nonzero, pairwise non-parallel vectors in $\R^3$ that realise every edge of a graph as an orthogonality (non-edges unconstrained).
\item[Obstruction.] A graph with no embedding. Any graph that contains it also has no embedding.
\item[Basis structure.] The hypergraph of bases (triangles) of an orthogonality graph. It is linear, its collinearity graph is squarefree (two vectors are adjacent when they share a basis), and it has no other triangles.
\item[Canonical augmentation.] McKay's method for generating combinatorial objects with exactly one representative per isomorphism class.
\item[Lex-leader constraint.] A clause set that removes assignments lexicographically smaller than their image under a symmetry. At least the lexicographically largest assignment of each orbit stays.
\item[CEGAR.] Counterexample-guided refinement: solve, check the model, and add a clause that excludes the model if it is spurious.
\item[Cube-and-conquer.] A method that splits a SAT instance into cubes (partial assignments). The cubes cover all cases, and the solver solves each cube separately.
\item[DRAT / LRAT.] Proof formats for SAT refutations. LRAT adds hints so that a small verified checker can validate the proof quickly.
\item[Positivstellensatz-style certificate.] A polynomial identity which shows that a system of real polynomial equations has no solution. Exact arithmetic checks the identity.
\end{description}

\FloatBarrier
\section{The statement}
\label{app:spec}
The definitions of \texttt{Spec.lean} (35 lines with imports and comments), followed by the final theorem:
\begin{lstlisting}
abbrev V3 := EuclideanSpace (*$\mathbb{R}$*) (Fin 3)
def NonPar (x y : V3) : Prop := LinearIndependent (*$\mathbb{R}$*) ![x, y]
def Admissible (V : Finset V3) : Prop := (*$\forall$*) x (*$\in$*) V, (*$\forall$*) y (*$\in$*) V, x (*$\neq$*) y (*$\to$*) NonPar x y
def Orth (x y : V3) : Prop := inner (*$\mathbb{R}$*) x y = 0
def VColouring (V : Finset V3) (f : V3 (*$\to$*) Bool) : Prop :=
  ((*$\forall$*) x (*$\in$*) V, (*$\forall$*) y (*$\in$*) V, Orth x y (*$\to$*) (*$\neg$*) (f x = true (*$\wedge$*) f y = true)) (*$\wedge$*)
  ((*$\forall$*) x (*$\in$*) V, (*$\forall$*) y (*$\in$*) V, (*$\forall$*) z (*$\in$*) V, x (*$\neq$*) y (*$\to$*) y (*$\neq$*) z (*$\to$*) x (*$\neq$*) z (*$\to$*)
      Orth x y (*$\to$*) Orth y z (*$\to$*) Orth x z (*$\to$*)
      f x = true (*$\vee$*) f y = true (*$\vee$*) f z = true)
def IsKS (V : Finset V3) : Prop := Admissible V (*$\wedge$*) (*$\neg$*) (*$\exists$*) f, VColouring V f
def NoKSUpTo (N : Nat) : Prop := (*$\forall$*) V : Finset V3, V.card (*$\le$*) N (*$\to$*) (*$\neg$*) IsKS V

theorem noKS_of_certified_extended (M N : (*$\mathbb{N}$*)) (lib ext : Array Obs)
    (hlib : (*$\forall$*) H (*$\in$*) (lib ++ ext).toList, (*$\exists$*) c, obsCheck H c = true)
    (hlow : (*$\forall$*) n, 1 (*$\le$*) n (*$\to$*) n (*$\le$*) M (*$\to$*) OrderCertified lib n)
    (hhigh : (*$\forall$*) n, M < n (*$\to$*) n (*$\le$*) N (*$\to$*) OrderCertified (lib ++ ext) n) : NoKSUpTo N
\end{lstlisting}
We apply the final theorem with $M=23$, $N=24$, \lean{lib}$\,=L^\ast$, and \lean{ext} set to the two new obstructions.

\FloatBarrier
\section{Certificate checkers}
\label{app:checkers}
\begin{itemize}
\item \lean{enumCheckF} checks a level-by-level witness that the structure list is complete up to isomorphism (\lean{enumComplete\_of}). For every candidate new basis, either a local test rejects it or a given permutation maps it onto a listed structure. \lean{enumCheck\_of\_structs} splits the check by structure, so the check runs in parallel with bounded memory.
\item \lean{checkSym} works on one structure. It builds the structure's CNF inside Lean. It checks the witness of every learned clause (a colouring, or an obstruction index with an injective map). It checks that each symmetry generator is an automorphism. It runs Lean's verified LRAT checker on CaDiCaL's proof. \lean{checkSym\_sound} proves that the lex-leader constraints preserve satisfiability. \lean{checkSymSplit} adds cube splits, with a coverage proof and one LRAT proof per cube.
\item \lean{obsCheck} checks real-algebraic non-embeddability certificates. One triangle is fixed to $e_1,e_2,e_3$. The other vertices are placed by cross products or in projective normal form, which gives a complete case split. Each case closes by polynomial identities $\sum_i m_i Z_i = \prod|u\times w|^2\sum_j s_j h_j^2$ with $s_j>0$. The certificate-generation scripts construct the identities with Singular, exact $LDL^\top$ factorisations, or a rounded SDP~\citep{goulart2024}.
\item \lean{LibExtend.lean} proves that every accepted check stays accepted when entries are appended to the obstruction library. Thus order 24 could use two new obstructions without a new check of orders 1--23.
\end{itemize}

\FloatBarrier
\section{Proof outline of the two-basis theorem}
\label{app:tb}
\begin{theorem}[two-basis case; computer-assisted]\label{thm:tb}
No KS set in $\R^3$ with 20 vectors has every vector in at most two of its bases.
\end{theorem}

Let $M$ be the graph whose nodes are the bases and whose edges are the vectors shared by two bases.
\begin{itemize}
\item Two bases that share two vectors are equal. Three bases that pairwise share a vector force those three vectors to be mutually orthogonal. The three vectors then form a fourth basis, and some vector lies in three bases. A 4-cycle of bases forces two parallel vectors. So $M$ is simple, with girth at least 5 and maximum degree 3.
\item Counting gives $n=3b-|E(M)|$, and the covering lemma gives $9\le b\le13$ at $n=20$. (Covering lemma: every KS set in $\R^3$ has at least 9 bases. For a unit $z$, the vectors $v$ with $(\hat v\cdot z)^2>\frac12$ are pairwise non-orthogonal. They meet every basis unless $z$ lies in the corner region of that basis, about 12.1\% of the sphere. With at most 8 bases some $z$ lies in no corner region, which gives a 010-colouring.)
\item \texttt{geng} (nauty~\citep{mckay2014}) lists exactly 330 such $M$. For each $M$, a SAT search over the extra edges with colouring and obstruction CEGAR is unsatisfiable. The 330 searches take 21\,s in total.
\end{itemize}
This theorem is not certified in Lean: it relies on \texttt{geng}, CaDiCaL and Z3-proved obstructions.

\FloatBarrier
\section{Order 24: full accounting}
\label{app:o24}
\begin{table}[!htbp]
\caption{The certified order-24 run on the laptop. \cpu\ is user+sys. $^{a}$Cube tree of the 8-disjoint-bases structure, built on a 4-core cloud machine. The Lean-side total for order 24 is 167{,}884 \cpu-seconds (46.6\,h, rows are rounded).}
\label{tab:o24}
\centering\footnotesize
\setlength{\tabcolsep}{4pt}
\resizebox{\linewidth}{!}{\begin{tabular}{lrrl}
\toprule
stage & wall & \cpu\ (s) & result\\
\midrule
1. enumeration (untrusted) & 1{,}373\,s & 1{,}373 & 2{,}200{,}310 partial structures, 19 levels\\
2. Lean enumeration check (2 parts) & 9{,}186\,s & 17{,}434 & accepted; 1{,}748{,}733 final structures\\
3. certificates (untrusted) & --- & --- & all but 67 closed by one LRAT proof each\\
3. cube trees, 65 hard structures & 8.8\,h$^{a}$ & --- & 45{,}870 cubes\\
4. Lean per-structure check (35 parts) & 7.8\,h & 123{,}271 & 1{,}748{,}729 accepted, 0 failed, 4 missing\\
4. Lean check, 8-disjoint-bases structure (12 parts) & 1.4\,h & 27{,}105 & 38{,}380 cubes and coverage accepted\\
4. Lean check, 3 remaining structures & 105\,s & 73 & accepted\\
\bottomrule
\end{tabular}}
\end{table}

\begin{figure}[!htbp]
\centering
\includegraphics[width=.9\linewidth]{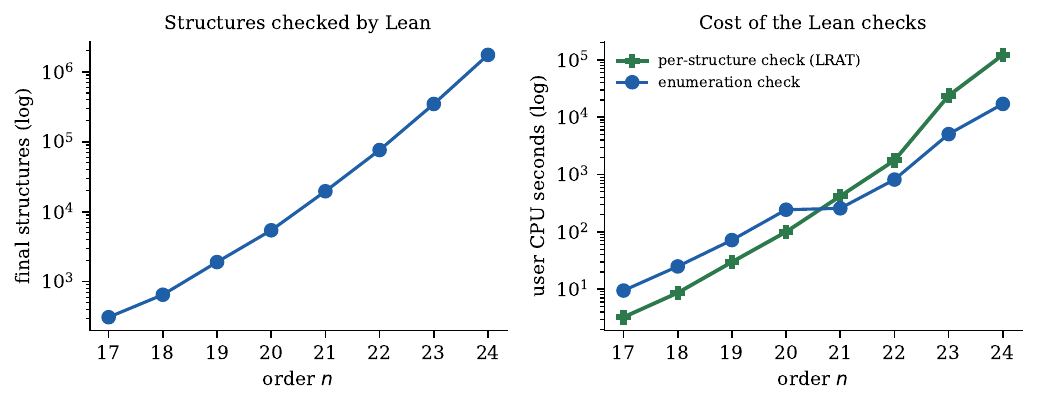}
\caption{Left: final structures per order. Right: user \cpu\ of the Lean checks. The per-structure check grows about $3$--$4\times$ per order through order 22. It then grows $13\times$ and $5\times$ at orders 23 and 24, where hard structures need cube splits.}
\label{fig:cost}
\end{figure}

The 35 per-structure parts checked 1{,}748{,}729 structures with no failure. The checker reported the other four as missing, not failed. The four parts that contain them (0, 1, 6 and 24) exited with an error status. A separate run then accepted each of the four:
\begin{itemize}
\item the 8-disjoint-bases structure, in 12 cube parts with coverage in part 0,
\item the two structures that need $L^{\ast\ast}$,
\item one hard structure whose certificate block was unreadable in the combined certificate file.
\end{itemize}
The supplementary structure identities are exactly the four missing primary identities (zero-based final-list indices 0, 1, 818341 and 1724404). Thus the accepted primary and supplementary sets are disjoint and their union is the full final list. A supplementary run checks one structure of its part and therefore reports the rest of that part as missing. Coverage rests on this identity accounting, not on the exit status of any single run. The 12 cube-check parts cover all 38{,}380 cube indices, and part 0 also accepts the Boolean coverage check. The two enumeration parts account for all 2{,}200{,}310 rows across 19 levels. Library checks cover every base-library index 0--1104 and the two appended indices 1105 and 1106. These are statements about the retained verification record, with the execution and provenance assumptions of Section~\ref{sec:cert}. The repository's coverage audit (\texttt{scripts/audit\_coverage.py}, output in \texttt{audit/coverage.json}) reproduces this accounting from the published inputs and logs.
Peak memory was 1.79\,GB per process. Over orders 1--24, Lean checked 2{,}200{,}309 final structures with no failure. This is exactly one fewer than the 2{,}200{,}310 rows of the order-24 enumeration. That enumeration has one row for every structure on at most 24 vectors, plus the empty structure at level~0. The same relation holds at every order in Table~\ref{tab:perorder}. The per-structure checks took about 41 \cpu-hours (user) and the enumeration checks 6.6, or 57.2 hours with system time (Figure~\ref{fig:cost}, Table~\ref{tab:perorder}).

\FloatBarrier
\section{Lean blueprint, and the verification obligations at each rung}
\label{app:blueprint}
Figure~\ref{fig:blueprint} shows the conditional Lean soundness theorems used to establish \lean{NoKSUpTo 24}. Every green node is a Lean theorem with no \lean{sorry}. The grey node is the LRAT checker that ships with Lean, and Lean proves that this checker is sound. Blue nodes are certificate data produced by untrusted tools. Only a checker reads each blue node (dashed arrows), and the soundness theorem of that checker sits directly above it. The top conclusion additionally relies on compiled checker execution and assembly of the accepted records, as stated in Section~\ref{sec:cert}.
The proof has the same shape for every order $n$. \lean{noKS\_of\_comb} inducts on the number of vectors. \lean{reduce} converts a minimum KS set into a reduced graph whose vertices all lie in triangles and which contains no library graph. \lean{OrderCertified lib n} excludes every such graph on $n$ vertices.
From order to order, only the certificate data supplied to \lean{OrderCertified} changes: the enumeration witness $E_n$ and the per-structure certificates $C_n$. All orders share the obstruction library.

Table~\ref{tab:perorder} gives the certificate data for every verified order separately. Order 0 needs no certificate (\lean{combOrder\_zero}).
For each order, \lean{enumCheckF} accepted the enumeration witness. \lean{checkSym} accepted every final structure or, for structures split into cubes, \lean{checkSymSplit} accepted it. No check failed.
The checkers check orders 1--23 against the library $L^\ast$ (1{,}105 graphs). They check order 24 against $L^{\ast\ast}$, which appends two new obstructions. \lean{orderCertified\_append} lifts orders 1--23 to $L^{\ast\ast}$ without a new run. \lean{obsCheck} accepted certificates for all 1{,}107 library graphs.

\begin{figure}[!htbp]
\centering
\begin{tikzpicture}[x=1cm,y=1cm,
  lab/.style={font=\scriptsize, anchor=east, inner sep=1pt},
  hd/.style={font=\scriptsize, rotate=50, anchor=west, inner sep=1pt},
  keep/.style={fill=black!30, minimum size=2.0mm, inner sep=0pt},
  new/.style={fill=agentd, draw=black, line width=0.3pt, minimum size=2.9mm, inner sep=0pt},
  gone/.style={font=\scriptsize, text=red!65!black, inner sep=0pt},
  ckeep/.style={fill=trustd, minimum size=2.0mm, inner sep=0pt},
  cnew/.style={fill=trustd, draw=black, line width=0.3pt, minimum size=2.9mm, inner sep=0pt}]
\def\rh{0.34}\def\cw{0.62}\def\cx{5.55}
\newcommand{\obrow}[5]{%
  \pgfmathsetmacro{\yy}{-#1*\rh}
  \draw[black!10] (-0.15,\yy) -- (\cx+0.3,\yy);
  \node[lab] at (-0.3,\yy) {#2};
  \foreach \n in {17,...,24} {
    \pgfmathsetmacro{\xx}{(\n-17)*\cw}
    \pgfmathtruncatemacro{\isnew}{(\n==#3) && (#3>17)}
    \pgfmathtruncatemacro{\iskeep}{(\n>=#3) && (\n<=#4) && (\isnew==0)}
    \pgfmathtruncatemacro{\isgone}{\n==#4+1}
    \ifnum\isnew=1 \node[new] at (\xx,\yy) {};\fi
    \ifnum\iskeep=1 \node[keep] at (\xx,\yy) {};\fi
    \ifnum\isgone=1 \node[gone] at (\xx,\yy) {$\times$};\fi
  }
  \ifcase#5 \or \node[ckeep] at (\cx,\yy) {};\or \node[cnew] at (\cx,\yy) {};\or \node[gone] at (\cx,\yy) {$\times$};\fi
}
\node[lab] at (-0.3,0.42) {rung (order $n$)};
\foreach \n/\t in {17/{orderly generation, colouring CEGAR},18/{cube-and-conquer},19/{obstruction library},20/{two-basis theorem},21/{\textbf{structure-first search}},22/{symmetry breaking, C++ solver},23/{lazy constraints, new obstructions},24/{exact transversals, shortcuts$^\dagger$}} {
  \pgfmathsetmacro{\xx}{(\n-17)*\cw}
  \node[font=\scriptsize] at (\xx,0.42) {\n};
  \node[hd] at (\xx-0.05,0.68) {\t};
  \draw[black!10] (\xx,0.22) -- (\xx,-22*\rh-0.18);
}
\node[font=\scriptsize] at (\cx,0.42) {1--24};
\node[hd, text=trustd] at (\cx-0.05,0.68) {\textbf{certified pipeline}};
\draw[black!10] (\cx,0.22) -- (\cx,-22*\rh-0.18);
\draw[black!45, dashed] (4.95,0.6) -- (4.95,-22*\rh-0.18);
\obrow{0}{statement, minimality reductions, LRAT checker}{17}{24}{1}
\obrow{1}{colouring clauses and their witnesses}{17}{24}{1}
\obrow{2}{polynomial non-embeddability certificates}{17}{24}{1}
\obrow{3}{orderly generation keeps the canonical graph}{17}{20}{0}
\obrow{4}{whole-graph encoding is sound}{17}{20}{0}
\obrow{5}{UNSAT proof for the whole order}{17}{20}{0}
\obrow{6}{prefix cubes cover the search space}{18}{20}{0}
\obrow{7}{obstruction library, cut soundness}{19}{24}{1}
\obrow{8}{witness for each obstruction cut}{19}{24}{1}
\obrow{9}{list of basis graphs from \texttt{geng}}{20}{20}{0}
\obrow{10}{two-basis theorem}{20}{20}{0}
\obrow{11}{vertex-0 dominance}{20}{20}{0}
\obrow{12}{covering lemma (at least 9 bases)}{20}{24}{3}
\obrow{13}{basis structures and completions}{21}{24}{1}
\obrow{14}{structure enumeration is complete}{21}{24}{1}
\obrow{15}{per-structure encoding is sound}{21}{24}{1}
\obrow{16}{UNSAT proof for each structure}{21}{24}{1}
\obrow{17}{lex-leader symmetry breaking is sound}{22}{24}{1}
\obrow{18}{certificates for obstructions found in the run}{23}{24}{1}
\obrow{19}{edge-minimality argument$^\dagger$}{24}{24}{3}
\obrow{20}{witnesses for solver-free rejections$^\dagger$}{24}{24}{3}
\obrow{21}{cube splits with a coverage proof}{99}{0}{2}
\obrow{22}{library-extension lemma}{99}{0}{2}
\foreach \r in {2,6,8,12,18,20} { \draw[black!40] (-6.2,-\r*\rh-0.5*\rh) -- (\cx+0.3,-\r*\rh-0.5*\rh); }
\def\ly{-8.15}
\node[new] at (-6.0,\ly) {}; \node[lab, anchor=west] at (-5.8,\ly) {new at this rung};
\node[keep] at (-3.5,\ly) {}; \node[lab, anchor=west] at (-3.3,\ly) {still required};
\node[gone] at (-1.5,\ly) {$\times$}; \node[lab, anchor=west] at (-1.3,\ly) {no longer required};
\node[ckeep] at (1.3,\ly) {}; \node[lab, anchor=west] at (1.5,\ly) {established};
\node[cnew] at (3.2,\ly) {}; \node[lab, anchor=west] at (3.4,\ly) {added for certification};
\end{tikzpicture}
\caption{Verification obligations at each rung. Columns 17--24 show what a certification of the search pipeline at that rung must establish. These columns are an analysis, not certified runs. Order 17 is the starting set. Orange: an obligation that the technique of that rung adds. Grey: an obligation that stays. $\times$: an obligation that is no longer necessary. The last column shows the certified pipeline for orders 1--24. Green: established by a Lean proof or an accepted check. Green with a border: added for certification. The certified pipeline does not use the covering lemma or the $\dagger$ shortcuts of Table~\ref{tab:ladder}. It keeps the structures with at most 8 bases and checks them directly. At order 24 these include the structure with 8 disjoint bases.}
\label{fig:obligations}
\end{figure}
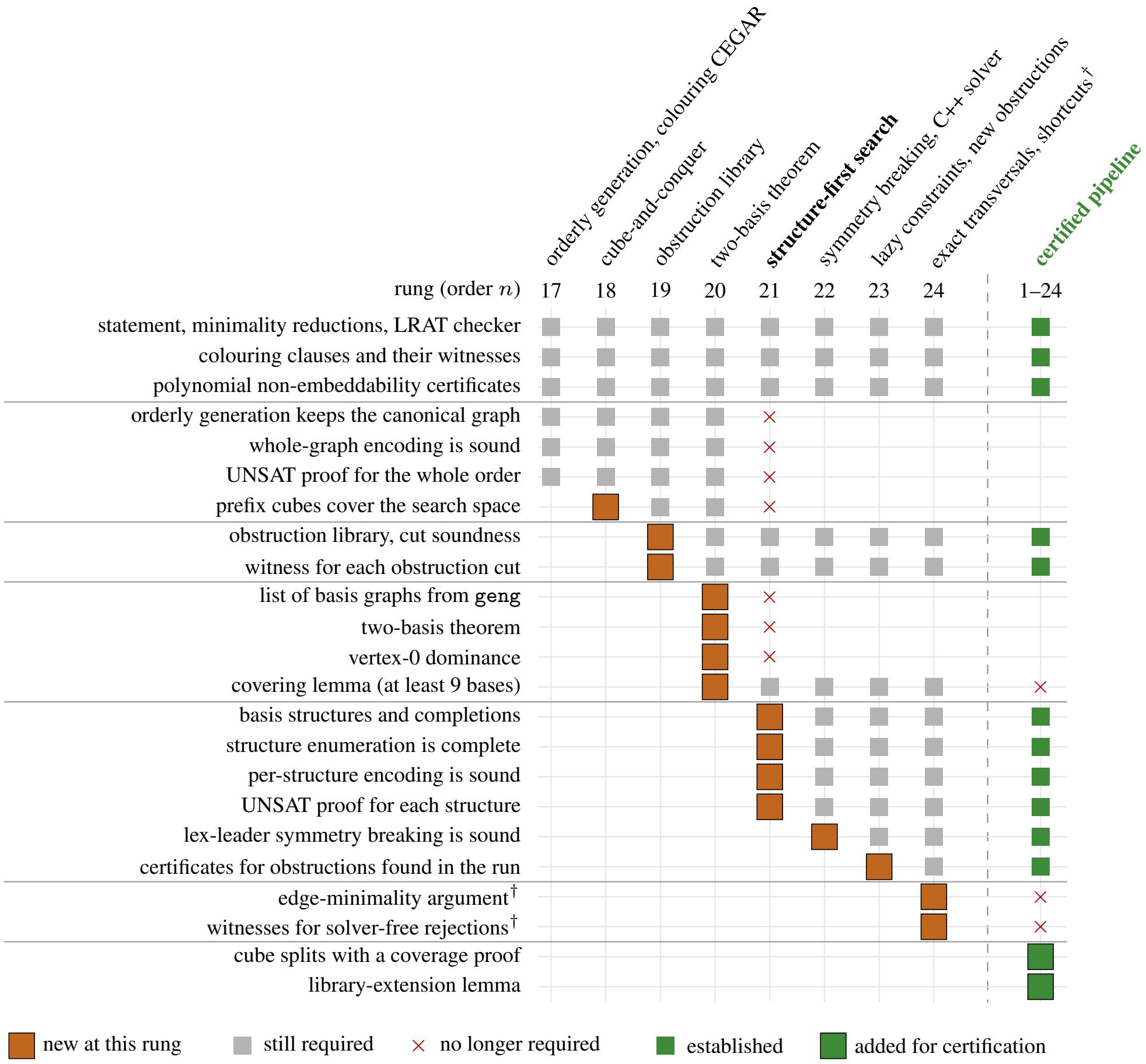

Figure~\ref{fig:obligations} shows the facts that a certification of the search pipeline at each rung must establish. Each new technique changes this set. The structure-first pipeline at order 21 replaces the two-basis proof branch with an enumeration of basis structures and SAT searches over their completions. Thus it no longer needs the two-basis theorem, the \texttt{geng} list of basis graphs, or the vertex-0 dominance restriction. It also removes the obligations of orderly generation and of the whole-order UNSAT proof. It adds obligations for the structure enumeration and for each structure. The two shortcuts at order 24 add obligations that are argued only on paper. At orders 21--24 the search still uses the covering lemma to skip structures with at most 8 bases. The certified pipeline keeps these structures and does not use the shortcuts. Thus it does not have these obligations. Columns 17--24 are an analysis of the search pipelines. Only the last column describes checks that were run.

\begin{table}[!htbp]
\caption{Every verified order. ``Structures'' and ``witnesses'' count all levels of the enumeration certificate. ``Final'' is the number of structures on exactly $n$ vectors. Sources: the per-order run logs. The acceptance records are in \texttt{logs/} of the public repository. The order-24 per-structure time sums the 35 parts and the four separately checked structures (Appendix~\ref{app:o24}).}
\label{tab:perorder}
\centering\scriptsize
\setlength{\tabcolsep}{2.4pt}
\begin{tabular}{rrrrrr@{\hspace{9pt}}lrrrr}
\toprule
 & \multicolumn{4}{c}{enumeration (\lean{enumCheckF})} & \multicolumn{2}{c}{per-structure checks} & & \multicolumn{2}{c}{Lean user \cpu\ (s)} & \\
\cmidrule(lr){2-5}\cmidrule(lr){6-7}\cmidrule(lr){9-10}
$n$ & levels & structures & witnesses & final & \lean{checkSym} & \lean{checkSymSplit} (cubes) & library & enum. & checks & failed\\
\midrule
1 & 1 & 1 & 0 & 0 & 0 & 0 & $L^\ast$ & 0 & 0.01 & 0\\
2 & 1 & 1 & 0 & 0 & 0 & 0 & $L^\ast$ & 0 & 0.01 & 0\\
3 & 2 & 2 & 1 & 1 & 1 & 0 & $L^\ast$ & 0 & 0.01 & 0\\
4 & 2 & 2 & 4 & 0 & 0 & 0 & $L^\ast$ & 0 & 0.01 & 0\\
5 & 3 & 3 & 13 & 1 & 1 & 0 & $L^\ast$ & 0 & 0.01 & 0\\
6 & 3 & 4 & 30 & 1 & 1 & 0 & $L^\ast$ & 0 & 0.01 & 0\\
7 & 4 & 6 & 71 & 2 & 2 & 0 & $L^\ast$ & 0 & 0.01 & 0\\
8 & 4 & 7 & 136 & 1 & 1 & 0 & $L^\ast$ & 0 & 0.01 & 0\\
9 & 5 & 12 & 285 & 5 & 5 & 0 & $L^\ast$ & 0.01 & 0.01 & 0\\
10 & 6 & 16 & 494 & 4 & 4 & 0 & $L^\ast$ & 0.02 & 0.02 & 0\\
11 & 6 & 25 & 923 & 9 & 9 & 0 & $L^\ast$ & 0.06 & 0.03 & 0\\
12 & 7 & 35 & 1{,}658 & 10 & 10 & 0 & $L^\ast$ & 0.13 & 0.05 & 0\\
13 & 8 & 60 & 3{,}012 & 25 & 25 & 0 & $L^\ast$ & 0.32 & 0.12 & 0\\
14 & 9 & 91 & 5{,}244 & 31 & 31 & 0 & $L^\ast$ & 0.65 & 0.19 & 0\\
15 & 11 & 169 & 9{,}998 & 78 & 78 & 0 & $L^\ast$ & 1.60 & 0.52 & 0\\
16 & 11 & 288 & 18{,}464 & 119 & 119 & 0 & $L^\ast$ & 3.72 & 1.09 & 0\\
17 & 12 & 598 & 36{,}524 & 310 & 310 & 0 & $L^\ast$ & 9.49 & 3.25 & 0\\
18 & 13 & 1{,}248 & 75{,}089 & 650 & 650 & 0 & $L^\ast$ & 25 & 8.66 & 0\\
19 & 14 & 3{,}153 & 169{,}659 & 1{,}905 & 1{,}905 & 0 & $L^\ast$ & 72 & 30 & 0\\
20 & 15 & 8{,}576 & 416{,}167 & 5{,}423 & 5{,}423 & 0 & $L^\ast$ & 244 & 100 & 0\\
21 & 16 & 28{,}196 & 1{,}178{,}952 & 19{,}620 & 19{,}620 & 0 & $L^\ast$ & 258 & 421 & 0\\
22 & 17 & 104{,}388 & 3{,}826{,}507 & 76{,}192 & 76{,}192 & 0 & $L^\ast$ & 819 & 1{,}772 & 0\\
23 & 18 & 451{,}577 & 14{,}705{,}862 & 347{,}189 & 347{,}188 & 1 (1{,}665) & $L^\ast$ & 5{,}083 & 23{,}839 & 0\\
24 & 19 & 2{,}200{,}310 & 65{,}833{,}376 & 1{,}748{,}733 & 1{,}748{,}668 & 65 (45{,}870) & $L^{\ast\ast}$ & 17{,}098 & 122{,}595 & 0\\
\midrule
total & & & & 2{,}200{,}309 & 2{,}200{,}243 & 66 (47{,}535) & & 23{,}616 & 148{,}771 & 0\\
\bottomrule
\end{tabular}
\end{table}

\FloatBarrier
\section{The pipeline at every order}
\label{app:pipelines}
Table~\ref{tab:components} lists the components of the pipeline that first solved each order. Table~\ref{tab:outcomes} shows how the structure-first pipeline closed the structures of each order. ``Orderly generation in SAT'' is the design of SAT+CAS~\citep{li2024} and SMS~\citep{kirchweger2023}: the solver builds the graph, and a canonicity check prunes non-canonical partial graphs. The agent used it only for orders 17--20 and never used the SMS tool itself.

\begin{table}[!htbp]
\caption{Components of the pipeline that first solved each order (Table~\ref{tab:ladder}). The structure-first pipeline later solved orders 17--20 again and certified orders 1--24. ``nauty'' symmetry breaking means isomorph-free enumeration by \texttt{genbg}. $^a$In the certified run only: the hardest structure at order 23 and the 65 hardest at order 24. The obstruction library is built from Z3 and Singular proofs, and \lean{obsCheck} checks it in Lean. $^b$An exact case analysis with Gröbner bases in SymPy showed that the single order-17 candidate is non-embeddable. Order 18 had no candidate.}
\label{tab:components}
\centering\scriptsize
\setlength{\tabcolsep}{3pt}
\begin{tabular}{c c c c c c c c c}
\toprule
$n$ & \mc{orderly gen.\\in SAT} & \mc{cube-and-\\conquer} & \mc{nauty enum.\\+ SAT per structure} & \mc{symmetry\\breaking} & \mc{obstruction\\library} & \mc{Z3\\\texttt{nlsat}} & \mc{Gröbner\\bases} & \mc{certified\\in Lean}\\
\midrule
17 & \checkmark & -- & -- & canonicity & -- & -- & SymPy$^b$ & \checkmark\\
18 & \checkmark & \checkmark & -- & canonicity & -- & -- & --$^b$ & \checkmark\\
19 & \checkmark & \checkmark & -- & canonicity & \checkmark & \checkmark & -- & \checkmark\\
20 & \checkmark & \checkmark & -- & canonicity & \checkmark & \checkmark & -- & \checkmark\\
21 & -- & -- & \checkmark & nauty & \checkmark & \checkmark & -- & \checkmark\\
22 & -- & -- & \checkmark & nauty + $\mathrm{Aut}(G_B)$ & \checkmark & \checkmark & -- & \checkmark\\
23 & -- & 1 structure$^a$ & \checkmark & nauty + $\mathrm{Aut}(G_B)$ & \checkmark & \checkmark & -- & \checkmark\\
24 & -- & 65 structures$^a$ & \checkmark & nauty + $\mathrm{Aut}(G_B)$ & \checkmark & \checkmark & -- & \checkmark\\
\bottomrule
\end{tabular}
\end{table}

\begin{table}[!htbp]
\caption{Recorded stage-2 outcomes through order 24 (\cpu: user+sys, stage 2 only for the certified rows, the whole run for the search rows). ``Certified'' rows produced certificates that Lean-proved checkers checked later. Exact-transversal contradictions can close a structure without a SAT call. ``Search'' rows describe the original discovery runs. Table~\ref{tab:perorder} gives the independent certificate coverage. $^a$The search skips structures with at most 8 bases, by the covering lemma (Appendix~\ref{app:tb}). The certified enumeration keeps them, which gives one more structure at orders 23 and 24. $^b$A first pass left two graphs unresolved. Z3 proved them non-embeddable, and they went into the library before this run. Table~\ref{tab:ladder} includes that discovery. $^c$Closed by the solver-free precheck$^\dagger$ without a SAT call.}
\label{tab:outcomes}
\centering\scriptsize
\setlength{\tabcolsep}{3pt}
\begin{tabular}{c l r r r r r r r}
\toprule
 & & & \multicolumn{4}{c}{structures closed by} & & \\
\cmidrule(lr){4-7}
$n$ & run & structures & \mc{no SAT call\\(trans./precheck)} & \mc{SAT:\\UNSAT} & \mc{obstruction\\library} & \mc{computer\\algebra} & SAT calls & \cpu\ (s)\\
\midrule
17 & certified & 310 & 174 & 135 & 1 & 0 & 136 & 0.3\\
18 & certified & 650 & 280 & 370 & 0 & 0 & 370 & 0.7\\
19 & certified & 1{,}905 & 881 & 1{,}020 & 4 & 0 & 1{,}024 & 2.4\\
20 & certified & 5{,}423 & 2{,}069 & 3{,}339 & 15 & 0 & 3{,}369 & 8.4\\
21 & certified & 19{,}620 & 6{,}868 & 12{,}651 & 101 & 0 & 13{,}134 & 9.8\\
22 & certified & 76{,}192 & 22{,}019 & 53{,}112 & 1{,}061 & 0 & 64{,}290 & 39.6\\
23 & search$^a$ & 347{,}188 & -- & 340{,}722 & 6{,}466 & 0$^b$ & 5{,}795{,}924 & 1{,}190$^b$\\
24 & search$^{a,\dagger}$ & 1{,}748{,}732 & 366{,}417$^c$ & 1{,}292{,}722 & 89{,}593 & 0 & 8{,}223{,}901 & 3{,}623\\
\bottomrule
\end{tabular}
\end{table}

\FloatBarrier
\section{Ablation sources and measured dead ends}
\label{app:dead}
\begin{figure}[!htbp]
\centering
\includegraphics[width=.95\linewidth]{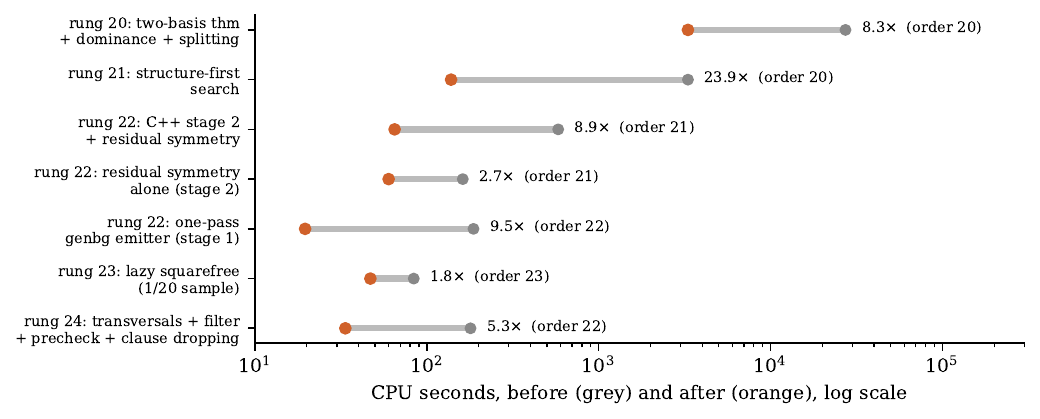}
\caption{Ablation: \cpu\ before (grey) and after (orange) each kept technique, measured on the instance shown. The structure-first search is the only change worth more than an order of magnitude on its own.}
\label{fig:ablation}
\end{figure}
\paragraph{Ablation (Figure~\ref{fig:ablation}).} All values are \cpu-seconds.
\begin{itemize}
\item Order 20: the full orderly search takes 27{,}298. With the two-basis theorem, dominance and splitting it takes 3{,}308. Structure-first takes 138.5.
\item Order 21: structure-first in Python takes 581.7. C++ with residual symmetry takes 65.0. Symmetry alone changes the solver stage from 162 to 60.
\item Order 22: enumeration per number of bases 187, one-pass enumeration 19.6.
\item Order 23: lazy squarefree constraints on a 1/20 sample, 84 to 47.
\item Order 22: the rung-24 bundle, 179.7 to 33.6.
\end{itemize}

\paragraph{Rejected ideas (25).}
\begin{itemize}
\item \emph{Orders 17--20 (12):} an edge-only partial-colourability propagator, lookahead cubes, vertex-criticality witnesses, partial-assignment canonicity, 3-cycle lex-leader, sharing learned or canonicity clauses between workers, pre-seeding all colouring clauses, an auxiliary-variable squarefree encoding ($+9\%$), ``forgettable'' canonicity and obstruction clauses ($+20\%$), LPT cube ordering, CaDiCaL retuning, a computational proof of the two-basis case inside the main solver.
\item \emph{Rung 21 (6), screened at $n=19$:} a vertex-0 degree clause ($+67\%$), the full obstruction library instead of the top 40 ($+29\%$), cube depth 11 or 13, at least $K$ vertices in three or more triangles ($+45\%$ to $4\times$), structure enumeration by adding vectors one at a time (stalled), structure enumeration by SAT (timed out).
\item \emph{Rung 22 (1):} a covering-based prune inside \texttt{genbg} (no effect).
\item \emph{Rung 23 (4):} fully lazy squarefree constraints (944M lazy clauses), eager threshold 2 (more SAT calls), phase 0 (no effect), shrinking by edge deletion (over-shrank to an unproven graph).
\item \emph{Rung 24 (2):} extra random group elements for lex-leader (slower), lazy edge-minimality cuts (never fire).
\end{itemize}

\end{document}